\pdfoutput=1

\documentclass[11pt]{article}

\usepackage[]{acl}

\usepackage{times}
\usepackage{latexsym}

\usepackage{graphicx}
\usepackage{booktabs}
\usepackage{subcaption}
\usepackage{adjustbox}
\usepackage{multirow}
\usepackage{amsmath,amsfonts,amssymb}
\usepackage{url}
\usepackage{xcolor}
\usepackage{todonotes}
\definecolor{red_v2}{RGB}{208, 2, 27}
\usepackage{subcaption}
\usepackage[normalem]{ulem}
\useunder{\uline}{\ul}{}

\usepackage[T1]{fontenc}

\usepackage[utf8]{inputenc}

\usepackage{microtype}

\title{Multi-Stage Prompting for Knowledgeable Dialogue Generation}

\author{Zihan Liu$^\ddagger$\thanks{~~This work was done when the first author was an intern at NVIDIA. Corresponding authors: Zihan Liu, Mostofa Patwary.}, Mostofa Patwary$^\mathsection$, Ryan Prenger$^\mathsection$, Shrimai Prabhumoye$^\mathsection$, \\ \textbf{Wei Ping$^\mathsection$, Mohammad Shoeybi$^\mathsection$, Bryan Catanzaro$^\mathsection$} \\ 
$^\ddagger$The Hong Kong University of Science and Technology, $^\mathsection$NVIDIA \\
\texttt{zihan.liu@connect.ust.hk}, \texttt{mpatwary@nvidia.com}
}

\begin{document}
\maketitle
\begin{abstract}
Existing knowledge-grounded dialogue systems typically use finetuned versions of a pretrained language model (LM) and large-scale knowledge bases. These models typically fail to generalize on topics outside of the knowledge base, and require maintaining separate potentially large checkpoints each time finetuning is needed. 
In this paper, we aim to address these limitations by leveraging the inherent knowledge stored in the pretrained LM as well as its powerful generation ability. 
We propose a multi-stage prompting approach to generate knowledgeable responses from a single pretrained LM.
We first prompt the LM to generate knowledge based on the dialogue context. Then, we further prompt it to generate responses based on the dialogue context and the previously generated knowledge.
Results show that our knowledge generator outperforms the state-of-the-art retrieval-based model by 5.8\% when combining knowledge relevance and correctness.
In addition, our multi-stage prompting outperforms the finetuning-based dialogue model in terms of response knowledgeability and engagement by up to 10\% and 5\%, respectively.
Furthermore, we scale our model up to 530 billion parameters and show that larger LMs improve the generation correctness score by up to 10\%, and response relevance, knowledgeability and engagement by up to 10\%. 
Our code is available at: \url{https://github.com/NVIDIA/Megatron-LM}.


\end{abstract}

\section{Introduction}

Dialogue systems face the problem of producing bland and generic outputs that are devoid of content~\cite{wolf2019transfertransfo,holtzman2019curious,ma2020survey}. Recent efforts have been made to address these concerns by grounding dialogue responses on a source of knowledge ~\cite{dinan2018wizard,zhou2018dataset,zhao2019low,santhanam2020local,prabhumoye2021focused}. Therefore, building a knowledgeable dialogue system has become one of the key milestone tasks in conversational research.

Current knowledge-grounded dialogue systems highly rely on a massive external knowledge corpus for a retrieval model to obtain relevant knowledge~\cite{dinan2018wizard,kim2019sequential,zhao2020knowledge}, which inevitably brings several limitations.
First, retrieval systems are constrained to the size and domains of the database, and they cannot generalize to out-of-domain topics that are not covered by the database. Second, retrieval from a massive corpus takes substantial resources. \citet{reimers2021curse} show that it is more difficult for the state-of-the-art retrieval model~\cite{karpukhin2020dense} to retrieve relevant knowledge when the size of the database increases. The larger database increases the chance that an irrelevant document is closer to the query embedding than the relevant document.




\begin{figure}[!t]
\centering
\resizebox{0.49\textwidth}{!}{
    \includegraphics{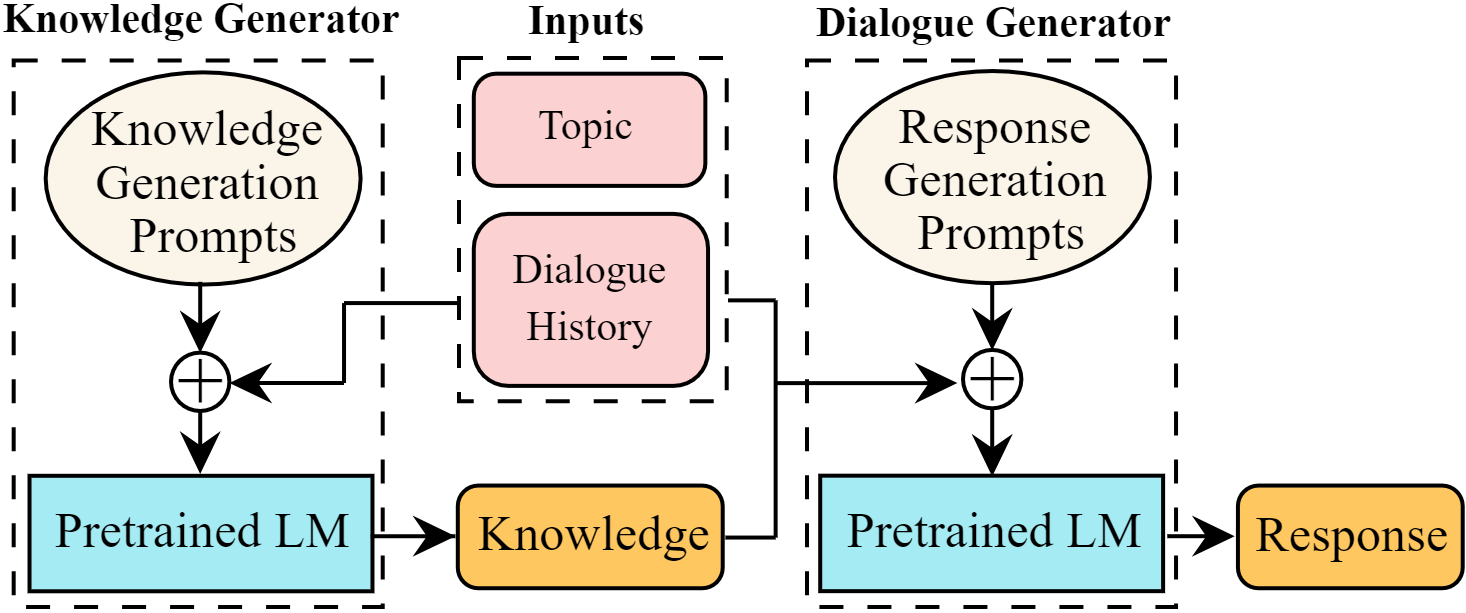}
}
\caption{Our proposed framework (MSDP) for the knowledgeable dialogue generation.}
\label{fig:framework}
\end{figure}

We aim to address these limitations by using a relatively small database and a pretrained language model (LM)~\cite{shoeybi2019megatron,brown2020language} as an additional source of knowledge to ground a dialogue system.
Since the LM inherently stores a variety of knowledge~\cite{petroni2019language}, it can help dialogue systems generalize to out-of-domain topics that are not explicitly present in the database. 
We propose a prompt-based approach to directly generate the context-relevant knowledge from the LM. Specifically, we select a few dialogue contexts and their associated knowledge from the database to be given as prompts to the LM for the knowledge generation. These samples are chosen such that the dialogue contexts are semantically closer to the current dialogue context.

Moreover, finetuning LMs, which current dialogue systems rely on, could lead to overfitting when the finetuning dataset is relatively small. Also, gigantic LMs like GPT-3~\cite{brown2020language} and Megatron-Turing NLG 530B~\cite{smith2022using}, may only be available through APIs. Hence, finetuning them on the dialogue task might not be a feasible solution.
To bypass the finetuning process, we propose to further prompt the LM to generate the response based on the dialogue context and previously generated knowledge. We select a few dialogue contexts and corresponding knowledge and responses to be given as prompts to the LM for the response generation. The samples are chosen such that their responses are knowledgeable and highly conditioned on the corresponding knowledge.

In summary, we present a novel \textbf{M}ulti-\textbf{S}tage \textbf{D}ialogue \textbf{P}rompting (MSDP) framework, which consists of a first-stage prompting for the knowledge generation and a second-stage prompting for the response generation. 
Our framework does not need any finetuning or updates to the pretrained weights of the LM, can generate relevant and factually correct knowledge, and is effective at producing knowledgeable and engaging responses.

Our contributions are summarized as follows:

\begin{itemize}
    \item We propose a novel multi-stage prompting framework for knowledgeable dialogue generation that only uses a single LM and does not require any finetuning.
    \item We show that for in-domain dialogue topics, our knowledge generator can outperform the state-of-the-art retrieval model by 5.8\% when combining relevance and correctness, and it can also better generalize to out-of-domain topics by a 6.4 F1-score improvement.
    \item We show that MSDP can outperform the finetuning-based dialogue model for response  knowledgeability and engagement by up to 10\% and 5\%, respectively.
    \item We scale our technique up to a 530-billion-parameter LM and demonstrate that larger LMs improve the generation correctness score by up to 10\%, and response relevance, knowledgeability and engagement by up to 10\%.
\end{itemize}

\section{Framework}
Our proposed multi-stage dialogue prompting (MSDP) framework is illustrated in Figure~\ref{fig:framework}. It consists of a knowledge generator and a dialogue generator, both using the same pretrained LM.
The knowledge generator produces relevant knowledge to the input topic and dialogue history, while the dialogue generator generates engaging and knowledgeable responses based on the dialogue context and the generated knowledge.

We denote the input topic as $t$, the input dialogue history as $h$, the last dialogue turn as $h^*$, and a database of samples as $D$. Each data sample in $D$ is denoted by $d_i$, and consists of a topic $t_i$, a dialogue history $h_i$ with the last turn as $h^*_i$, corresponding knowledge $k_i$, and a response $r_i$.

\begin{figure}[t!]
\centering
\resizebox{0.49\textwidth}{!}{
    \includegraphics{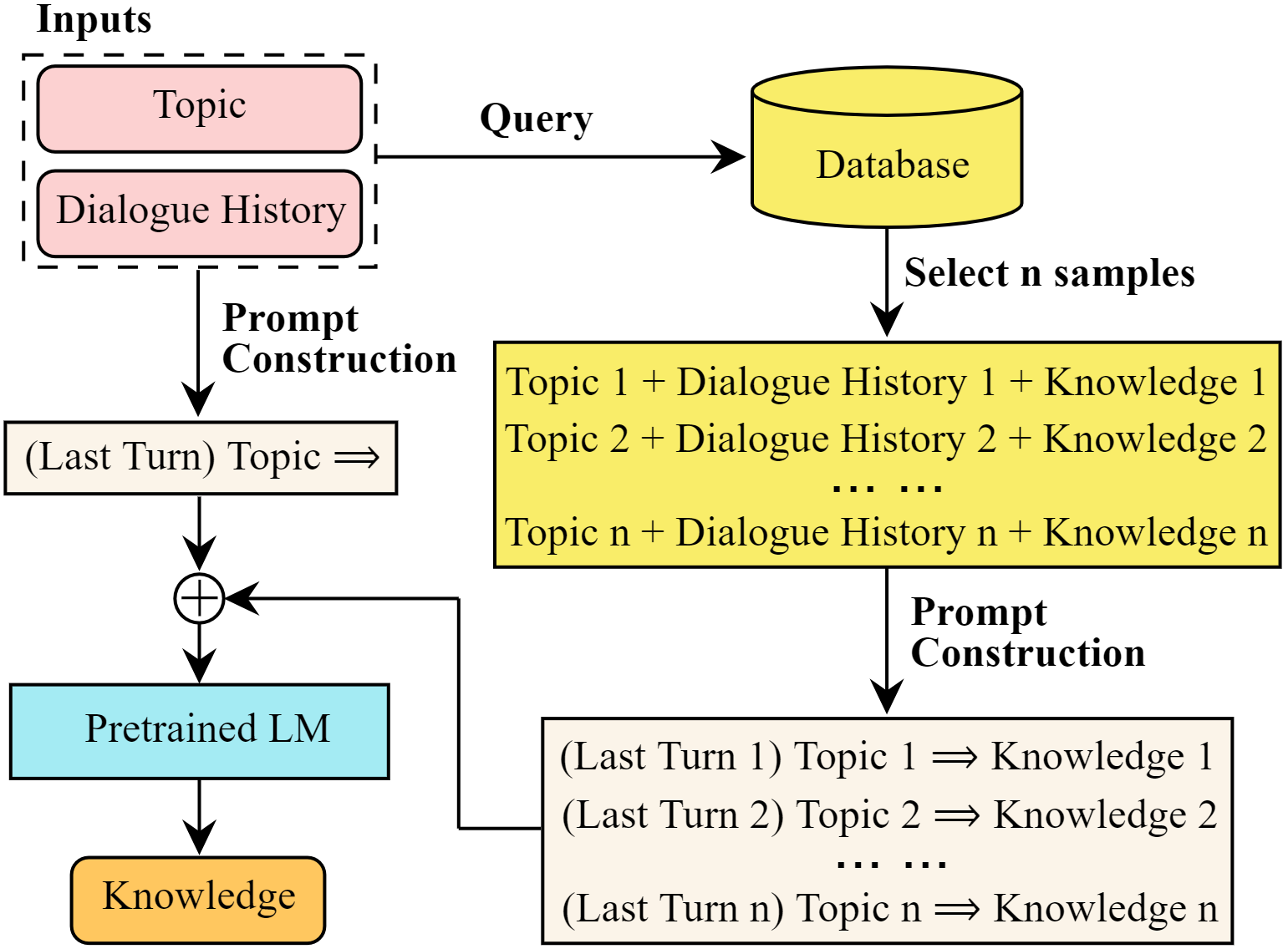}
}
\caption{Prompting for the knowledge generation.}
\label{fig:promptKnowledge}
\end{figure}

\subsection{Knowledge Generator}
To bypass the dependence on a large-scale knowledge base, we propose a prompt-based knowledge generation approach, which uses a relatively small database (about 70K samples) and a pretrained LM to generate context-relevant knowledge.
As shown in Figure~\ref{fig:promptKnowledge}, the knowledge generation consists of sample selection and knowledge generation.

\paragraph{Sample Selection}

We hypothesize that selecting appropriate samples as prompts is the key to generating high-quality knowledge sentences.
Intuitively, leveraging the knowledge from similar topics or dialogue context can help the LM to generate contextually relevant and factually correct knowledge sentences.
Hence, we propose a query-based sample selection method, which aims to search similar samples from $D$ based on the input query ($q$).
To ensure that the selected examples are relevant to the query, we utilize a pretrained sentence encoder ($SE$)~\cite{devlin2019bert,karpukhin2020dense} to obtain the representations for the query and each data sample ($d_i$) in $D$. Then, we calculate the similarity between the query and each sample using the dot product of their representations:
\begin{equation*}
    Sim(q, d_i) = SE(t+h)^\intercal \cdot SE(t_i + h_i),
\end{equation*}
where the input of the $SE$ is a concatenation of the topic and dialogue history. Finally, we select $n$ samples that have the highest similarity scores to $q$. This selection process can be done efficiently since the database is relatively small.

\paragraph{Knowledge Generation}
\label{knowledge_gen}
Inspired by the few shot approach in~\citet{brown2020language}, feeding the pretrained LM with suitable and intuitive prompts can allow it to generate relevant content. 
The template of the constructed prompts is illustrated in Figure~\ref{fig:promptKnowledge}.
Concretely, the prompt for the $i^{th}$ sample ($\textit{prompt}_i$, $i \in [1,n]$) is ``$(h^*_i)~t_i \Rightarrow k_i$''\footnote{For example, ( I love pizza ) Pizza $\Rightarrow$ Pizza is a traditional Italian dish typically topped with tomato sauce and cheese.}, and the prompt for the current dialogue context ($\textit{prompt}_{curr}$) is ``$(h^*)~t \Rightarrow$'', where we use the symbol ``$\Rightarrow$'' to guide the LM for knowledge generation.
We only use the last dialogue turn to construct prompts because the previous turns are mostly not relevant to the knowledge, and adding redundant information could lead to negative effects for knowledge generation. 
Given that $k_i$ usually has a closer connection with $t_i$ than $h^*_i$, we put $k_i$ closer to $t_i$ than $h^*_i$ in the prompts.
Finally, we concatenate the constructed prompts using ``\verb!\n!'' and feed them into the LM to generate the knowledge:
\begin{equation*}
    k' = \mathcal{LM}(\textit{prompt}_1 \verb!\n! ~ ... ~ \textit{prompt}_n \verb!\n! ~ \textit{prompt}_{curr})
\end{equation*}
where $k'$ denotes the generated knowledge for the input. Since ``\verb!\n!'' is used to separate the prompts, the model will start generating ``\verb!\n!'' followed by another random example after finishing the knowledge generation. 
Hence, we consider the generated sentence before ``\verb!\n!'' as $k'$.

\subsection{Dialogue Generator}

\begin{figure}[t!]
\centering
\resizebox{0.49\textwidth}{!}{
    \includegraphics{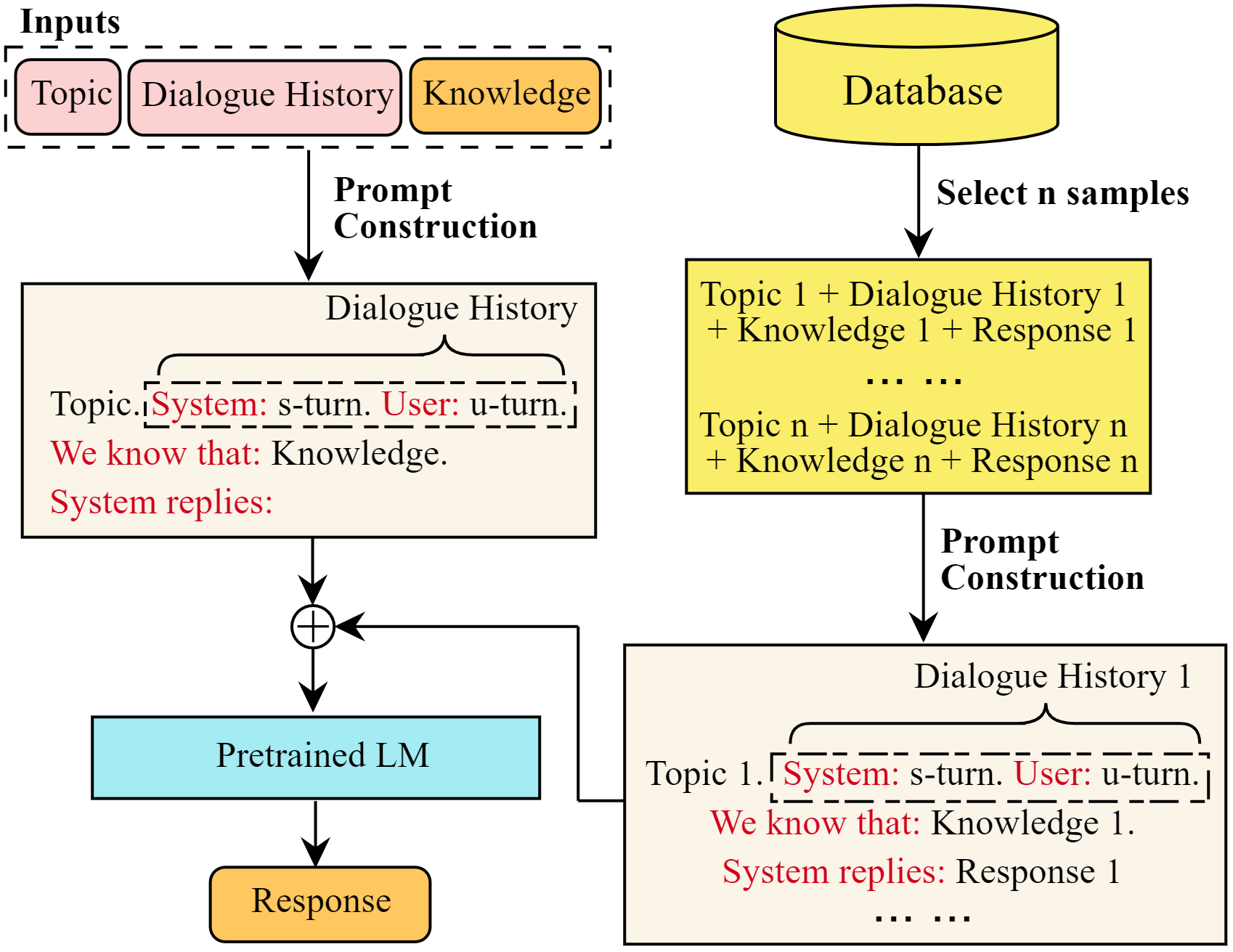}
}
\caption{Prompting for the dialogue response generation. We use comprehensive words (denoted in \textcolor{red_v2}{red} color) to connect the dialogue history, knowledge and response for the prompt construction.}
\label{fig:promptResponse}
\end{figure}

The architecture of our proposed dialogue generator is illustrated in Figure~\ref{fig:promptResponse}.
Finetuning a LM could lead to overfitting when the finetuning dataset is relatively small. In addition, since usually one can only access to the gigantic LMs, like GPT-3~\cite{brown2020language} and Megatron-Turing NLG 530B~\cite{smith2022using} using only APIs, finetuning them might not be a feasible solution.
Therefore, we propose to circumvent the finetuning by prompting the pretrained LM for the response generation, which requires only a few dialogue examples.
To generate knowledgeable and engaging responses, we focus on how to select samples and how to effectively prompt the LM for the response generation.


\paragraph{Sample Selection}
\label{sampleselection}
One of the essential skills for the knowledgeable dialogue model is to effectively leverage the knowledge produced in the first stage, in order to make the generated responses knowledgeable.
Considering that we can provide the LM with only a few dialogue samples, it could be difficult for it to learn how to generate a response based on the knowledge unless there are strong connections between the dialogue response and knowledge in the samples that we provide.
Hence, we focus on selecting the samples in which the responses are knowledgeable and highly conditioned on the corresponding ground truth knowledge.
Concretely, for each example in the database, we calculate how much ground truth knowledge accounts for the dialogue response by using the word overlap ratio. Then, we filter out the examples where the ratio is lower than 0.6 (this number is decided based on a hyper-parameter search among \{0.4, 0.5, 0.6, 0.7, 0.8\}).
Also having responses be too knowledgeable could make it less engaging.
Therefore, we also filter out the examples where the ratio is higher than 0.9 since we expect the response to contain information other than the knowledge.
After the filtering, to ensure that our approach does not depend on the dialogue context, we randomly select $n$ samples from the rest of the dialogue examples. These selected $n$ samples will be later constructed as prompts and used for the response generation.

\paragraph{Response Generation}
Aside from the ability to leverage the generated knowledge, another essential skill for the dialogue model is to have the ability to chat based on the dialogue context.
To equip our model with this skill, we focus on constructing intuitive prompts for the selected examples and feed them into the LM. 
The constructed prompts for the selected examples and inputs are illustrated in Figure~\ref{fig:promptResponse}.
For prompts from the selected examples, we use ``System:'' and ``User:'' to connect different turns in the dialogue history, and ``We know that:'' and ``System replies:'' are used to introduce the knowledge and response, respectively. 
For prompts from the current conversation (i.e., inputs), we follow the same template except that we keep the response empty for the pretrained LM to generate.

After the prompt construction, we concatenate the prompts for selected samples and the inputs using ``\verb!\n!'', and then feed them into the pretrained LM to generate the response. 
Similar to what we have described in Section \ref{knowledge_gen}, we consider the generated sentence before ``\verb!\n!'' as the response.

\section{Experimental Setup}
\subsection{Datasets}
We evaluate our model using two knowledge-grounded dialogue datasets: Wizard of Wikipedia (WoW)~\cite{dinan2018wizard} and Wizard of Internet (WoI)~\cite{komeili2021internet}.

WoW uses Wikipedia as the knowledge base and covers a wide range of topics (1365 in total). 
Its test dataset is split into two subsets: test seen and test unseen. 
Each data sample has a chosen topic, a dialogue history, a ground truth knowledge sentence, and a corresponding dialogue response. 
The dialogue topics in the test seen subset appear in the training dataset, while the topics in the test unseen subset do not.
Different from WoW, the collection of WoI is grounded on the whole Internet, which covers a wider range of topics than Wikipedia.

In the experiments, we only use the training set of WoW (as the database) for the sample selection of our prompting framework.
All the models (our model and baselines) \textit{do not use any training sample from WoI}, and we directly evaluate them on the test set of WoI.
The motivation for doing this is to test how well our model can generalize to the unseen scenario where topics do not exist in the database. The topics in the WoW test unseen set do not exist in the database, and only 5.76\% of topics in the WoI test set exist in the database. We calculate the 13-gram overlap~\cite{brown2020language} between the knowledge used in WoI test set and the database, and find the overlap is as little as 0.39\%. 

\begin{table*}[!htb]
\renewcommand{\arraystretch}{1.1}
\centering
\begin{adjustbox}{width={0.8\textwidth},totalheight={\textheight},keepaspectratio}
\begin{tabular}{l|cccc|cccc|cccc}
\toprule
\multicolumn{1}{l|}{\multirow{2}{*}{\textbf{Models}}} & \multicolumn{4}{c|}{\textbf{Wizard of Wikipedia (Seen)}} & \multicolumn{4}{c|}{\textbf{Wizard of Wikipedia (Unseen)}} & \multicolumn{4}{c}{\textbf{Wizard of Internet}} \\
\multicolumn{1}{c|}{}                                 & \textbf{B}  & \textbf{M}  & \textbf{R-L}  & \textbf{F1}  & \textbf{B}   & \textbf{M}   & \textbf{R-L}  & \textbf{F1}  & \textbf{B}   & \textbf{M}   & \textbf{R-L}   & \textbf{F1}  \\ \midrule
DPR (seen)   & 18.32       & 12.82       & 21.91         & 24.86        & 8.09         & 6.80         & 12.04         & 13.71        & 2.37         & 3.90         & 5.73           & 7.03         \\
DPR (wiki)       & 9.95        & 9.27        & 15.11         & 18.42        & 10.06         & 9.80         & 15.46         & 18.24        & 3.49         & 5.36         & 7.35           & 9.16         \\
FKG         & 21.08       & 14.61       & 25.57         & 27.83        & 9.01         & 8.26         & 15.61         & 16.07        & 3.45         & 4.69         & 6.55           & 7.14         \\ \midrule
MSDP-KG$^\dagger$      & \textbf{23.68}       & \textbf{15.93}       & \textbf{27.88}         & \textbf{31.55}        & \textbf{11.54}         & \textbf{10.53}        & \textbf{19.05}         & \textbf{20.15}        & \textbf{5.20}         & \textbf{7.38}         & \textbf{10.47}          & \textbf{11.12}       \\ \bottomrule
\end{tabular}
\end{adjustbox}
\caption{Results of automatic metrics for the knowledge generation/retrieval models across three datasets. B, M, and R-L denote the averaged BLEU, METEOR, and ROUGE-L, respectively. DPR (seen) can only access the knowledge in the training dataset of WoW,
while DPR (wiki) can access all the knowledge in Wikipedia. $^\dagger$We use ``-KG'' to denote the knowledge generation part of MSDP (same for the following tables). Both FKG and MSDP-KG use a 126m LM to match the size of DPR, which is based on a 110m LM.}
\label{tab:knowledgeGenResults}
\end{table*}

\begin{table*}[!htb]
\renewcommand{\arraystretch}{1.1}
\centering
\begin{adjustbox}{width={0.999\textwidth},totalheight={\textheight},keepaspectratio}
\begin{tabular}{l|ccc|ccc|ccc}
\toprule
\multirow{2}{*}{\textbf{Models}} & \multicolumn{3}{c|}{\textbf{Wizard of Wikipedia (Seen)}} & \multicolumn{3}{c|}{\textbf{Wizard of Wikipedia (Unseen)}} & \multicolumn{3}{c}{\textbf{Wizard of Internet}} \\ 
& \textbf{Relevance}      & \textbf{Correctness}     & \textbf{Combination}     & \textbf{Relevance}      & \textbf{Correctness}      & \textbf{Combination}      & \textbf{Relevance}   & \textbf{Correctness}  & \textbf{Combination}  \\ \midrule
DPR (110m)              & 3.39           & \textbf{4.00}            & 3.39         & 3.38           & \textbf{4.00}             & 3.38          & 2.79        & \textbf{4.00}         & 2.79      \\
MSDP-KG (126m)              & 3.76*          & 3.71            & 3.59*        & 3.80*          & 3.19             & 3.12          & 3.60*       & 2.93         & 2.83      \\
MSDP-KG (357m)              & 3.79*          & 3.80            & 3.69*        & 3.84*          & 3.56*            & 3.47          & 3.74*       & 3.29*        & 3.21*     \\
MSDP-KG (1.3b)              & 3.81*          & 3.90*           & 3.72*        & 3.89*          & 3.72*            & 3.62*         & 3.77*       & 3.51*        & 3.38*     \\
MSDP-KG (530b)              & \textbf{3.88}*          & 3.96*           & \textbf{3.84}*        & \textbf{3.92}*          & 3.94*            & \textbf{3.87}*         & \textbf{3.81}*       & 3.84*        & \textbf{3.70}*    \\ \bottomrule
\end{tabular}
\end{adjustbox}
\caption{Human evaluations for the knowledge generation/retrieval models. We compare MSDP-KG with DPR (seen) on the WoW (seen) dataset, and DPR (wiki) on the WoW (unseen) and WoI datasets. We directly use a score of 4 to rate the correctness of the knowledge retrieved by DPR since all knowledge in the database is correct.
For relevance and combination, we conduct a t-test between MSDP-KG and DPR. For the correctness, we conduct a t-test between MSDP-KG (357m-530b) and MSDP-KG (126m). * denotes the result is significant at p < 0.05.}
\label{tab:humanevalKnowledgeGen}
\end{table*}

\subsection{Baselines for Knowledge Generation}
\paragraph{DPR}
DPR, Dense Passage Retriever~\cite{karpukhin2020dense}, is the state-of-the-art retrieval model.
To make DPR fit into the dialogue scenario, we finetune it on the training dataset of WoW.
Concretely, it is finetuned to map the dialogue context (topic and dialogue history pair) and corresponding ground truth knowledge into similar vector space.\footnote{The details of this finetuning is placed in Appendix~\ref{appendixF}.}

\paragraph{FKG}
FKG denotes the finetuning-based knowledge generation.
We use the training dataset of WoW to finetune the LM. Concretely, the input is a concatenation of a topic and dialogue history, and the LM is finetuned to generate relevant knowledge.
We use FKG as a baseline to compare the performance of the knowledge generation between the prompt-based and finetuning-based methods.

\subsection{Baselines for Response Generation}
\paragraph{PPLM} PPLM denotes the plug and play language model~\cite{dathathri2019plug}.
We choose it as a baseline because our MSDP can be considered as using topics to control the LM to generate responses, and PPLM, which does not need finetuning either, can be also used to control LMs for topic-relevant generation.
We follow~\citet{madotto2020plug} and use dialoGPT~\cite{zhang2020dialogpt} for PPLM to enable the response generation. We use ConceptNet~\cite{speer2017conceptnet} to produce topic-relevant bag-of-words for the response generation.


\paragraph{FCM w/ DPR} 
FCM denotes the finetuning-based conversational model. 
We use the training dataset of WoW to finetune the LM. 
This baseline has the same pipeline as that of our MSDP. Instead of doing prompting, it uses DPR for producing the knowledge and FCM to generate a response.



\paragraph{FCM w/ FKG} ~ This baseline follows the same setting as FCM w/ DPR, except that we use FKG instead of DPR to produce knowledge.

Note that we do not compare our model with ~\citet{kim2019sequential,zhao2019low,zhao2020knowledge,zhan2021augmenting} that incorporate the information of the ground truth knowledge for the response generation since our model does not leverage such information (more details are available in Appendix~\ref{appendixG}). In addition, given that our model does not need any fine-tuning and uses only 20 samples as prompts for the response generation, FCM w/ DPR and FCM w/ FKG make them strong baselines for our model to compare with, since they were finetuned on the entire training dataset.



\subsection{Automatic Evaluation}
For evaluating both knowledge generation and response generation, we follow previous works~\cite{rashkin2019towards,dinan2018wizard,prabhumoye2021focused} to evaluate the generated sentences against the reference sentences on averaged BLEU (an average of BLEU-1,2,3,4)~\cite{papineni2002bleu}, ROUGE-L~\cite{lin2004rouge}, METEOR~\cite{denkowski2011meteor}, and unigram F1. Additionally, we follow~\citet{komeili2021internet} to use knowledge F1 (KF1) to evaluate the knowledgeability of the response generation.

\begin{table*}[!htb]
\renewcommand{\arraystretch}{1.0}
\centering
\begin{adjustbox}{width={0.99\textwidth},totalheight={\textheight},keepaspectratio}
\begin{tabular}{l|ccccc|ccccc|ccccc}
\toprule
\multicolumn{1}{l|}{\multirow{2}{*}{\textbf{Models}}} & \multicolumn{5}{c|}{\textbf{Wizard of Wikipedia (Seen)}} & \multicolumn{5}{c|}{\textbf{Wizard of Wikipedia (Unseen)}} & \multicolumn{5}{c}{\textbf{Wizard of Internet}} \\
\multicolumn{1}{c|}{}   & \textbf{B}  & \textbf{M}  & \textbf{R-L}  & \textbf{F1} & \textbf{KF1} & \textbf{B}   & \textbf{M}   & \textbf{R-L}  & \textbf{F1} & \textbf{KF1}  & \textbf{B}   & \textbf{M}   & \textbf{R-L}   & \textbf{F1} & \textbf{KF1}  \\ \midrule
\multicolumn{1}{l|}{PPLM}   & 2.08   & 4.89   & 6.32   & 11.40    & 6.63   & 2.15   & 4.86   & 6.30   & 11.38    & 6.77   & 1.78   & 4.58   & 5.70   & 9.83    & 4.48   \\
\multicolumn{1}{l|}{FCM w/ DPR (seen)}   & 8.72   & 8.40   & 14.91   & 17.40    & 17.13   & 6.51   & 6.88   & 12.12   & 13.71    & 11.54   & 4.06   & 6.27   & 9.17    & 12.90    & 7.38   \\
\multicolumn{1}{l|}{FCM w/ DPR (wiki)}   & 7.36   & 7.63   & 13.65   & 16.00    & 13.80   & 6.98   & 7.43   & 13.33   & 15.46    & 13.38   & 4.47   & 6.65   & 9.65    & 13.52    & 7.78   \\
\multicolumn{1}{l|}{FCM w/ FKG}    & 8.97   & 8.67   & 15.36   & 18.31    & 18.85   & 6.73   & 7.19   & 12.97   & 14.68    & 12.59  & 4.75   & 6.56   & 9.72    & 13.71    & 7.89 \\ \midrule
\multicolumn{1}{l|}{FCM w/ MSDP-KG}     & \textbf{10.17}   & 9.34   & 16.00   & \textbf{19.45}    & 21.02   & 7.12   & 7.70   & 13.93   & \textbf{16.75}    & 13.96   & \textbf{4.80}   & 6.82   & \textbf{10.21}    & \textbf{14.39}    & 8.77   \\
\multicolumn{1}{l|}{MSDP}     & 9.97   & \textbf{9.95}   & \textbf{18.62}   & 17.57    & \textbf{22.95}   & \textbf{8.30}   & \textbf{8.65}   & \textbf{17.40}   & 16.00    & \textbf{16.57}   & 4.66   & \textbf{8.00}   & 9.80    & 14.09    & \textbf{9.67}   \\ \bottomrule
\end{tabular}
\end{adjustbox}
\caption{Results of automatic metrics for the knowledgeable conversational model. Both FKG and MSDP-KG (associated with FCM) use a 126m LM to match the size of DPR, which is based on a 110m LM. MSDP uses a 357m LM to match the size of FCM, which is also based on a 357m LM.}
\label{tab:dialogResults}
\end{table*}

\subsection{Human Evaluation}
\paragraph{Knowledge Generation}
For evaluating the quality of the knowledge generation, we use \textbf{relevance}, \textbf{correctness}, and a \textbf{combination} of the two metrics.
To evaluate the relevance, we provide annotators with the topic and dialogue, as well as the model-produced knowledge, and ask them to rate how relevant the knowledge is to the topic and dialogue on a scale from 1 to 4, where 1 means not relevant at all, 2 is only a little relevant, 3 is somewhat relevant, and 4 is highly relevant.
To evaluate the correctness, we provide the annotators with the topic and the model-generated knowledge, and ask them to rate how correct the knowledge is on a scale from 1 to 4, where 1 is not correct at all, 2 is less than half is correct, 3 is half and more than half is correct, and 4 is all correct. 

In addition, given that the overall quality of the knowledge depends on both relevance and correctness, we calculate a combination score based on the minimum between the relevance and correctness for each evaluated sample:
\begin{equation*}
    \text{combination} = \textit{min}(\text{relevance}, \text{correctness}).
\end{equation*}
We use minimum instead of average or maximum because both relevance and correctness are indispensable for the quality of the knowledge.

\paragraph{Response Generation}
For evaluating the quality of the response generation, we use \textbf{relevance}, \textbf{engagement}, and \textbf{knowledgeability}.
To evaluate the relevance, we provide the annotators with a topic and dialogue history, as well as a pair of generated responses from two models and ask them to choose which is more relevant to both topic and dialogue history.
For engagement and knowledgeability, we provide the annotators with the same samples as for relevance, and ask them to choose which is more engaging and knowledgeable, respectively. 
For all these metrics, we let annotators choose a tie when the quality is comparable.\footnote{We put the human evaluation setup in the Appendix~\ref{appendixE}.}


\subsection{Training Details}
The LMs used for our MSDP model, and baselines FKG and FCM are GPT-style~\cite{brown2020language} models and are pretrained using the toolkit in~\citet{shoeybi2019megatron}.
PPLM uses dialoGPT-medium, which has 355 million parameters (355m). The LM in FCM has 357m parameters, and DPR consists of two encoders (question encoder and passage encoder) with a size of 110m parameters each.
To test how different model sizes affect the performance, we evaluate our methods with 126m, 357m, 1.3 billion (1.3b), and 530 billion (530b) parameters LMs.
For the sample selections, we choose 10 samples for the prompting in the knowledge generation, and 20 samples for the prompting in the response generation.
To ensure a fair comparison, we select the top-1 knowledge from the DPR model, and we use deterministic greedy search for the generation of LM.
We use the question encoder of DPR as the sentence encoder in the sample selection of the knowledge generation. Note that this sentence encoder can be replaced with any pretrained model (e.g., BERT~\cite{devlin2019bert}), and as shown in Section \ref{ablationstudy}, there is only a marginal difference between using BERT and DPR's question encoder (about 0.5 F1 for the dialogue response generation).

\section{Results}
In this section, we compare our framework with baselines for the knowledge and response generation. Then, we conduct ablation studies to further analyze the effectiveness of our framework.

\subsection{Knowledge Generation}
We first analyze how DPR performs when different sizes of databases are available.
From Table~\ref{tab:knowledgeGenResults}, we can see that in the WoW (seen) scenario, DPR (seen) can retrieve generally better knowledge compared to DPR (wiki) since the corpus size for DPR (wiki) is much larger.  This further confirms that larger database makes retrieval of relevant information more difficult DPR as shown in~\citet{reimers2021curse}.
However, DPR (seen) cannot work in the unseen scenarios (WoW (unseen) and WoI) due to the absence of a topic-relevant knowledge base.
Compared to DPR, FKG achieves better results when the topics are covered in the training dataset (WoW (seen)), while its generalization ability to unseen topics is relatively limited since we can see that DPR (wiki) has better performance than FKG in WoW (unseen) and WoI.
Our approach, MSDP-KG, demonstrates a powerful generalization ability to unseen topics, which leads to better results across the three datasets compared to all the baselines.

To evaluate the generation quality, we compare MSDP-KG with DPR using human evaluation, and the results are shown in Table~\ref{tab:humanevalKnowledgeGen}. 
We find that MSDP-KG (126m) can generate much more relevant knowledge compared to DPR (with more than 10\% improvement in the relevance score). 
In addition, MSDP-KG (126m) can produce generally correct knowledge in WoW (seen) since it can refer to the knowledge in similar topics, which leads to a better combination score than DPR (a 5.8\% improvement).
Meanwhile, its generation correctness is somewhat limited in WoW (unseen) and WoI, which can be attributed to the relatively small model size and the pretraining corpus.
We notice that MSDP-KG (126m) also achieves a better combination score in WoI due to a very significant improvement in the relevance score. 
This is because the knowledge base for DPR is limited in the Wikipedia domain, which lowers its generalization ability to a wider range of topics on the Internet.

Furthermore, we observe that larger LMs bring improvements on all metrics. 
MSDP-KG (357m) can outperform DPR in all datasets for the combination score. We find that larger LMs can also bring significant improvement on the correctness score (e.g., 357m improves over 126m by 11.5\% in WoW (unseen)). Moreover, MSDP-KG (530b) achieves a 3.94 correctness score for WoW (unseen), which means about 95\% of the generations are all correct.

\begin{table}[t!]
\renewcommand{\arraystretch}{1.0}
\centering
\begin{adjustbox}{width={0.49\textwidth},totalheight={\textheight},keepaspectratio}
\begin{tabular}{lcccr}
\toprule
\multicolumn{1}{l|}{\textbf{Model A}}    & \textbf{Rele.}         & \textbf{Enga.}      & \multicolumn{1}{c|}{\textbf{Know.}}  & \textbf{Model B}    \\ \bottomrule
\multicolumn{5}{c}{\textbf{\textit{Wizard of Wikipedia (Seen)}}}   \\ \midrule
\multicolumn{1}{l|}{M (357m)} & 41.5 - 40.0 & \textbf{43.7} - 38.5 & \multicolumn{1}{c|}{\textbf{50.4} - 37.8} & F (357m)  \\
\multicolumn{1}{l|}{M (1.3b)} & \textbf{48.9} - 40.0 & \textbf{47.8} - 37.8 & \multicolumn{1}{c|}{\textbf{47.8} - 35.6} & M (357m) \\
\multicolumn{1}{l|}{M (530b)} & \textbf{54.4} - 41.1 & \textbf{53.3} - 41.1 & \multicolumn{1}{c|}{\textbf{51.1} - 42.2} & M (1.3b) \\ \bottomrule
\multicolumn{5}{c}{\textbf{\textit{Wizard of Wikipedia (Unseen)}}}        \\ \midrule
\multicolumn{1}{l|}{M (357m)} & 39.3 - 40.0 & \textbf{46.7} - 43.0 & \multicolumn{1}{c|}{\textbf{48.9} - 37.8} & F (357m)  \\
\multicolumn{1}{l|}{M (1.3b)} & \textbf{50.0} - 38.9 & \textbf{51.1} - 41.1 & \multicolumn{1}{c|}{\textbf{46.7} - 41.1} & M (357m) \\
\multicolumn{1}{l|}{M (530b)} & \textbf{52.2} - 42.2 & \textbf{51.1} - 40.0 & \multicolumn{1}{c|}{\textbf{50.0} - 38.9} & M (1.3b) \\ \bottomrule
\multicolumn{5}{c}{\textbf{\textit{Wizard of Internet}}}         \\ \midrule
\multicolumn{1}{l|}{M (357m)} & 42.2 - 43.7 & 41.5 - 40.7 & \multicolumn{1}{c|}{\textbf{44.4} - 39.3} & F (357m)  \\
\multicolumn{1}{l|}{M (1.3b)} & \textbf{51.1} - 42.2 & \textbf{50.0} - 38.9 & \multicolumn{1}{c|}{\textbf{44.4} - 41.1} & M (357m) \\
\multicolumn{1}{l|}{M (530b)} & \textbf{54.4} - 38.9 & \textbf{52.2} - 42.2 & \multicolumn{1}{c|}{\textbf{56.7} - 38.9} & M (1.3b) \\ \bottomrule
\end{tabular}
\end{adjustbox}
\caption{Human evaluation results on the dialogue models in terms of relevance (Rele.), engagement (Enga.), and knowledgeability (Know.). M denotes the MSDP and F denotes the FCM w/ DPR (DPR (seen) for WoW (seen), and DPR (wiki) for WoW (unseen) and WoI). For each number pair, the left number denotes the win rate for model A and the right one for model B. Note that the numbers in each pair might not sum to 100 since the annotators can choose ``tie''.}
\label{tab:humanevalDialoGen}
\end{table}

\subsection{Response Generation}
The automatic metrics for conversational models are shown in Table~\ref{tab:dialogResults}. 
We notice that PPLM does not perform as well as the other models for this task since it does not explicitly use the relevant knowledge for the response generation.
For the FCM-based models, we find that a better knowledge generation leads to a performance improvement as does a better retrieval model.
``FCM w/ MSDP-KG'' outperforms baseline models. 
Interestingly, our MSDP also generally outperforms the FCM-based baselines on different automatic metrics, especially the KF1 score. 
For example, compared to ``FCM w/ DPR (wiki)'', MSDP has a 3.19 higher KF1 score in WoW (unseen) and a 1.89 higher KF1 score in WoI. 
This can be attributed to the sample selection for the response generation, which selects knowledgeable responses that are highly based on the knowledge sentence. 
We also observe that MSDP achieves comparable results to the ``FCM w/ MSDP-KG'', which further illustrates the effectiveness of our proposed framework.

The human evaluations from Table~\ref{tab:humanevalDialoGen} further confirms the effectiveness of MSDP. Compared to ``FCM w/ DPR'', MSDP can generate relevant responses, and more engaging and knowledgeable responses. 
For WoW (seen) and WoW (unseen), MSDP has more a than 10\% higher win rate in terms of knowledgeability, and about 3\% to 5\% higher win rate in terms of the engagement. Furthermore, we observe that larger LMs generally improve on response relevance, engagement, and knowledgeability by about 10\% win rate. We also discuss about how different prompt formats impact the responses in Appendix~\ref{appendixI}.

\paragraph{In-depth Analysis of Generated Responses}
We observe that the generated response tends to partially copy the generated knowledge. This is due to the fact that the generated response is highly conditioned on the corresponding ground truth knowledge-response pairs in the prompts, and similar patterns exist in those pairs~\footnote{We put some generation samples in Appendix~\ref{appendixD}.}.

To have an in-depth analysis about the response generation, we quantify the proportion of the knowledge in the generated responses, which we formulate as follows:
\begin{equation*}
    ratio_{knwl} = \frac{\texttt{\# \{overlap tokens\}}}{\texttt{\# \{response tokens\}}},
\end{equation*}
where \texttt{\# \{overlap tokens\}} denotes the number of overlap tokens between the generated knowledge and the generated response. \texttt{\# \{response tokens\}} denotes the number of tokens in the response.
The ratios for MSDP using 357m, 1.3b, and 530b parameters in the WoW (unseen) are 49.67\%, 46.11\%, and 44.19\%, respectively. This suggests that the response is not just simply copies of the knowledge, it also contains additional information to ensure the relevance and engagingness.
Moreover, in Appendix~\ref{appendixH}, we showcase some examples where the generated knowledge is not very relevant to the conversation, and our model could manage to generate coherent and engaging responses.

\begin{table}[!t]
\renewcommand{\arraystretch}{1.1}
\centering
\begin{adjustbox}{width={0.49\textwidth},totalheight={\textheight},keepaspectratio}
\begin{tabular}{l|cccc|cccc}
\toprule
\multirow{2}{*}{\textbf{Models}} & \multicolumn{4}{c|}{\textbf{WoW (Seen)}} & \multicolumn{4}{c}{\textbf{WoW (Unseen)}} \\   & \textbf{B}       & \textbf{M}       & \textbf{R-L}    & \textbf{F1}     & \textbf{B}       & \textbf{M}       & \textbf{R-L}     & \textbf{F1}     \\ \midrule
MSDP-KG              & \textbf{24.5}    & \textbf{16.4}    & \textbf{28.7}   & \textbf{33.2}   & \textbf{12.4}    & \textbf{11.1}    & \textbf{19.6}    & \textbf{22.0}   \\
\quad w/ BERT      & 23.1    & 15.5    & 27.3   & 31.1   & 12.1    & 10.5    & 19.0    & 21.2   \\
\quad w/ random    & 12.9    & 9.72    & 17.6   & 18.8   & 9.85    & 10.1    & 17.5    & 19.8   \\
\quad w/o topic    & 21.5    & 14.2    & 25.3   & 27.2   & 7.37    & 6.86    & 13.3    & 14.2   \\ \bottomrule
\end{tabular}
\end{adjustbox}
\caption{Ablation studies for the knowledge generation, in terms of the sentence encoder (w/ BERT), sample selection method (w/ random), and the importance of the input topic (w/o topic). The size of the LM is 357m.}
\label{tab:pkgAblationStudy}
\end{table}

\begin{table}[!t]
\renewcommand{\arraystretch}{1.1}
\centering
\begin{adjustbox}{width={0.38\textwidth},totalheight={\textheight},keepaspectratio}
\begin{tabular}{l|ccccc}
\toprule
\multirow{2}{*}{\textbf{Models}} & \multicolumn{5}{c}{\textbf{Wizard of Wikipedia (Unseen)}} \\
& \textbf{B}       & \textbf{M}       & \textbf{R-L}      & \textbf{F1}      & \textbf{KF1}     \\ \midrule
MSDP                    & \textbf{8.30}    & \textbf{8.65}    & \textbf{17.40}    & \textbf{16.00}   & \textbf{16.57}   \\
\quad w/ BERT            & 8.13    & 8.38    & 17.16    & 15.51   & 16.13   \\
\quad w/ random          & 5.56    & 6.50    & 16.48    & 14.32   & 13.13   \\
\quad w/o topic          & 6.32    & 7.17    & 15.70    & 13.06   & 11.77  \\ \bottomrule
\end{tabular}
\end{adjustbox}
\caption{Ablation studies for the response generation, in terms of the sentence encoder in the knowledge generation, sample selection method, and the importance of an input topic. The size of the LM is 357m.}
\label{tab:msdpAbaltionStudy}
\end{table}

\subsection{Ablation Studies}
\label{ablationstudy}

\paragraph{Sentence Encoder}
In the sample selection of the knowledge generation, we obtain the similarity based on the DPR's question encoder, and we investigate how effective the generation will be if we replace the question encoder with a simpler model, like BERT~\cite{devlin2019bert}. 
From Table~\ref{tab:pkgAblationStudy}, using BERT as the sentence encoder achieves comparable performance to using DPR's question encoder. 
Also, from Table~\ref{tab:msdpAbaltionStudy}, we can see that using BERT in MSDP-KG only slightly lowers the performance in the response generation. These results confirms the effectiveness of our proposed method.

\paragraph{Sample Selection}
We study the effectiveness of our sample selection methods in both knowledge generation and response generation by using the random selection as a comparison.
From Table~\ref{tab:pkgAblationStudy}, we can see that using randomly selected samples consistently decreases the performance in all metrics. Since the random selection does not leverage the information from the database, the performance drop is especially significant in WoW (seen).
In addition, from Table~\ref{tab:msdpAbaltionStudy}, ``MSDP'' significantly outperforms ``MSDP w/ random'' in all metrics, which confirms the effectiveness our proposed sample selection for the response generation.

\paragraph{Importance of Input Topic}
In our framework, a topic is a part of the input. 
To investigate the effectiveness of using a topic, we remove the input topic from the knowledge generation and response generation. 
As shown in both Table~\ref{tab:pkgAblationStudy} and Table~\ref{tab:msdpAbaltionStudy}, we can see that providing a topic in the input is important, especially for the unseen scenario, where we observe a 7 F1-score decrease for ``MSDP-KG w/o topic'' in WoW (unseen).

\begin{figure}
\centering
\begin{subfigure}{.47\textwidth}
  \centering
  \includegraphics[width=.99\linewidth]{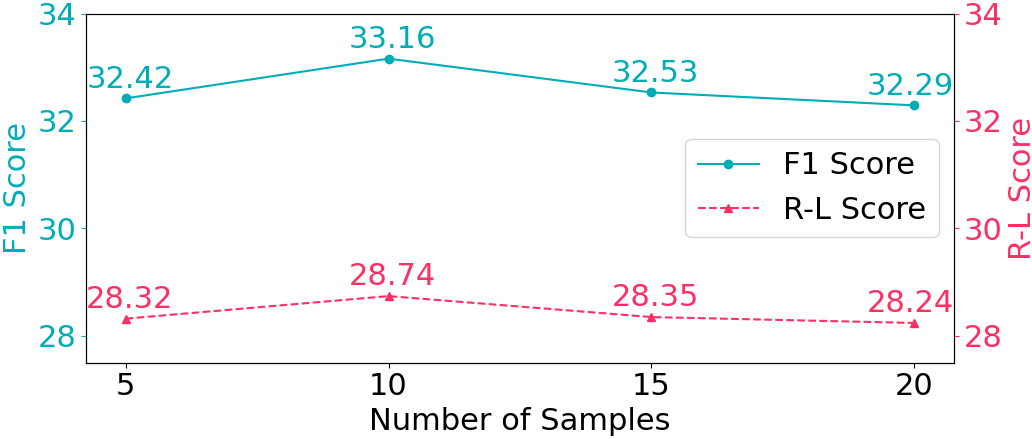}
\end{subfigure}
\begin{subfigure}{.47\textwidth}
  \centering
  \includegraphics[width=.99\linewidth]{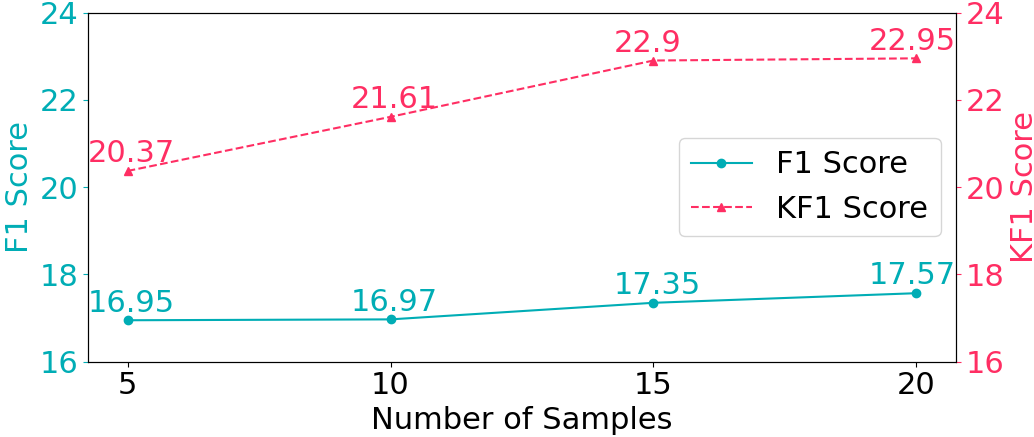}
\end{subfigure}
\caption{Effectiveness for different numbers of samples for the knowledge generation (top) and response generation (bottom). The size of the LM is 357m, and the results are from WoW (unseen).}
\label{fig:k_shot}
\end{figure}

\paragraph{Number of Samples for Prompting}
We further study how sample size affects the prompting performance. 
From Figure~\ref{fig:k_shot} (top), the number of samples will not significantly affect the knowledge generation. Interestingly, the performance of knowledge generation starts to slightly drop when sample size increases from 10. We conjecture that selecting too many samples might induce less similar samples to the input dialogue context, which could impact the performance negatively. As shown in Figure~\ref{fig:k_shot} (bottom), having more samples can slightly bring better responses. This is because, with more samples as references, the LM can better understand how to generate responses based on the given knowledge, which leads to a higher F1 and KF1 scores.\footnote{More ablation studies and results of automatic metrics for the model scaling are in the Appendix~\ref{appendixA}, \ref{appendixB}, and \ref{appendixC}.}

\paragraph{Multi-Stage Prompting vs. Single-Stage Prompting}
To further study the effectiveness of knowledge generator in our framework, we compare MSDP with single-stage dialogue prompting (SSDP). SSDP removes the stage of the knowledge generation, and directly uses the topic and the dialogue history to prompt the LM for the response generation. 
We keep the dialogue samples that are used to construct the response generation prompts the same for MSDP and SSDP. For the prompt design of SSDP, we simply remove the knowledge part (``We know that: \{Knowledge\}'') from the original one, due to the absence of the knowledge.
Table~\ref{tab:multistage-singlestage} illustrates the comparison between MSDP and SSDP. We find that MSDP remarkably outperforms SSDP across all metrics, especially for KF1. The results confirms that the stage of the knowledge generation in MSDP is highly important and indispensable.

\begin{table}[!t]
\renewcommand{\arraystretch}{1.1}
\centering
\begin{adjustbox}{width={0.49\textwidth},totalheight={\textheight},keepaspectratio}
\begin{tabular}{l|ccccc|ccccc}
\toprule
\multirow{2}{*}{\textbf{Models}} & \multicolumn{5}{c|}{\textbf{WoW (Seen)}} & \multicolumn{5}{c}{\textbf{WoW (Unseen)}} \\   & \textbf{B}       & \textbf{M}       & \textbf{R-L}    & \textbf{F1}  & \textbf{KF1}    & \textbf{B}       & \textbf{M}       & \textbf{R-L}     & \textbf{F1}  & \textbf{KF1}    \\ \midrule
SSDP        & 7.50    & 8.00    & 16.63   & 14.16   & 11.01    & 6.81    & 7.89   & 16.28  &  14.07 & 11.34 \\
MSDP      & \textbf{9.97}    & \textbf{9.95}    & \textbf{18.62}   & \textbf{17.57}   & \textbf{22.95}    & \textbf{8.30}    & \textbf{8.65}    & \textbf{17.40} & \textbf{16.00} & \textbf{16.57}   \\ \bottomrule
\end{tabular}
\end{adjustbox}
\caption{Comparisons between MSDP and SSDP.}
\label{tab:multistage-singlestage}
\vspace{-4mm}
\end{table}

\section{Related Work}
\subsection{Language Model Prompting}
Pretrained LMs are shown to possess commonsense knowledge~\cite{davison2019commonsense,bosselut2019comet,rajani2019explain,zhou2020evaluating}, and can be prompted to do cloze questions~\cite{petroni2019language,jiang2020can,brown2020language,shin2020eliciting,schick2021s,qin2021learning}, as well as many downstream natural language understanding and generation tasks, such as sentiment analysis, natural language inference, question answering, and text summarization~\cite{brown2020language,madotto2020language,zeng2021pangu,smith2022using,kumar2021reordering,shin2021constrained,wang2021want}.
\citet{li2021prefix} incorporated prompting and finetuning, and proposed prefix-tuning, which kept language model parameters frozen and optimized a small continuous task-specific vector for generation tasks. \citet{lester2021power} introduced prompt tuning, a simplification of prefix-tuning, and showed that prompt tuning became more competitive with scale.
Despite the extensive research having explored the LM prompting methods, little research has focused on directly generating context-relevant knowledge from LMs.

Recently, \citet{zheng2021exploring} and \citet{madotto2021few}, in concurrent works to ours, proposed to prompt LMs for the dialogue generation. Different from them, we focus on the knowledge-grounded scenario and propose a multi-stage prompting framework to leverage the inherent knowledge in LMs.




\subsection{Knowledge-grounded Dialogues}
Grounding dialogue responses based on a knowledge base ensures a knowledgeable and engaging response and is emerging as an important step in research of human-machine conversation~\cite{zhu2017flexible,ghazvininejad2018knowledge,dinan2018wizard,zhou2018dataset,kim2019sequential,moon2019opendialkg,zhao2019low,chen2020bridging,Li2020Zero,wu2020controllable,hedayatnia2020policy,zhan2021augmenting,prabhumoye2021focused,rashkin2021increasing,komeili2021internet}. 
\citet{kim2019sequential} proposed sequential knowledge transformer to boost the knowledge selection quality from the candidates, and improved the performance of the response generation. \citet{zhao2020knowledge} equipped the response generation defined by a pre-trained language model with a knowledge selection module, and jointly optimized them. 
Taking this further, \citet{komeili2021internet} extended the knowledge base to the whole Internet, which allowed a boarder coverage of the knowledge and more robust response generation quality.
Unlike the previous works, our proposed framework circumvents the need of LM finetuning and a massive knowledge base, which current models typically rely on.

\section{Conclusion}
We propose a novel multi-stage dialogue prompting framework which consists of a first-stage prompting for the knowledge generation and a second-stage prompting for the response generation. 
Both automatic metrics and human evaluations show that compared to the state-of-the-art retrieval-based model, our knowledge generator can generate better context-relevant knowledge for both in-domain and out-of-domain dialogue topics.
Moreover, our framework is able to produce more knowledgeable and engaging responses compared to the finetuning-based dialogue model. 
Additionally, we conduct comprehensive ablation studies to show the effectiveness of our proposed methods.
Furthermore, we scale the LM up to 530 billion parameters and demonstrate that larger LMs consistently improve the generation correctness, and response relevance, knowledgeability, and engagement.

\bibliography{anthology,custom}
\bibliographystyle{acl_natbib}

\appendix

\clearpage

\section{Perplexity-based Sample Selection} \label{appendixA}

\begin{figure}[!t]
\centering
\resizebox{0.49\textwidth}{!}{
    \includegraphics{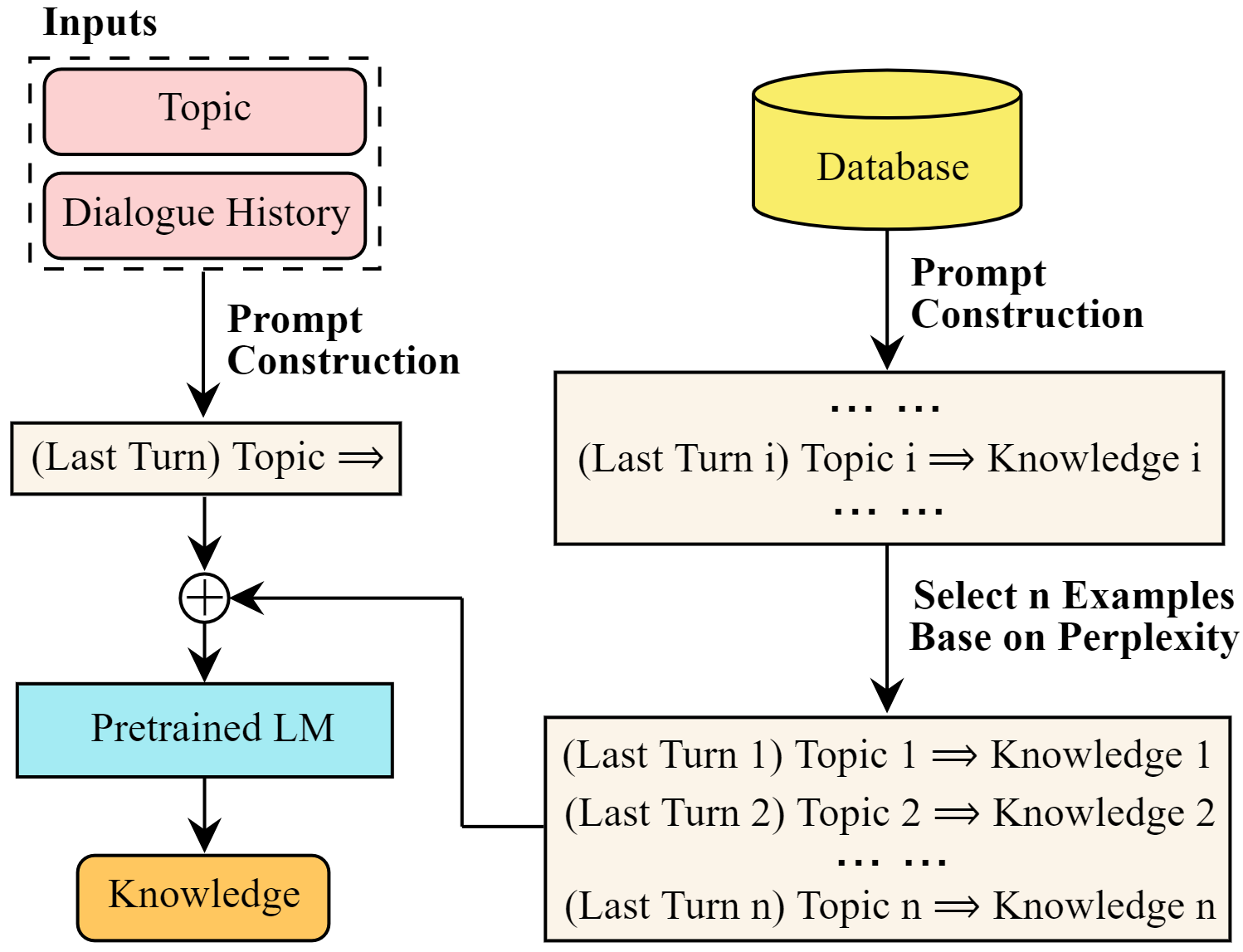}
}
\caption{Prompting for the knowledge generation using the perplexity-based sample selection.}
\label{fig:knowledgeGenPPLBased}
\end{figure}

We investigated another sample selection method (i.e., perplexity-based selection) for the knowledge generation. The knowledge generation using perplexity-based selection is depicted in Figure~\ref{fig:knowledgeGenPPLBased}. The details of this sample selection is described as follows. Note that we denote the sample selection method for the knowledge generation in the main paper (Section 2.1) as the query-based sample selection.

Instead of selecting samples based on the current conversation (i.e., query), perplexity-based method will complete the sample selection before the inference, and the selected examples can be used for all inputs (i.e, topic and dialogue history pairs).
Intuitively, using easy to understand prompts (instead of incomprehensible ones) enables the pretrained language models quickly comprehend the task and push it to generate the knowledge that is more topic-relevant and factually correct.
To find comprehensible prompts, we first perform the prompt construction\footnote{The prompt construction is the same as the query-based sample selection proposed in the main paper.} for each data example in the database. We then calculate the perplexity for each prompt using a GPT-2 model~\cite{radford2019language} and select top-$n$ prompts that have the lowest perplexities.\footnote{To ensure a fair comparison with the query-based sample selection in the main paper (Section 2.1), we choose top-10 samples for the perplexity-based sample selection.}

Compared to query-based selection, the prompts selected based on perplexities are less relevant to the test example, which could generally lead to a worse generation quality. However, its advantage is that we do not need to select samples from the database for every input. Technically, it needs only a few easy to understand samples (i.e., 10 samples) for prompting.


\section{Ablation Studies Results} \label{appendixB}
In the ablation study, we compare the query-based sample selection method (used in MSDP) and the perplexity-based sample selection method.
We also provide the automatic metrics for different model sizes. We denote the sample selection method for the knowledge generation in the main paper (Section 2.1) as the query-based selection. In the tables, we use ``ppl.'' to denote that the model is using the perplexity-based sample selection for the knowledge generation, and ``que.'' to denote that the model is using the query-based sample selection for the knowledge generation.

\begin{table}[]
\renewcommand{\arraystretch}{1.1}
\centering
\begin{adjustbox}{width={0.43\textwidth},totalheight={\textheight},keepaspectratio}
\begin{tabular}{lcccc}
\toprule
\multicolumn{1}{l|}{\textbf{Models}}           & \textbf{B} & \textbf{M} & \textbf{R-L} & \textbf{F1} \\ \bottomrule
\multicolumn{5}{c}{\textit{\textbf{Wizard of Wikipedia (Seen)}}}      \\ \midrule
\multicolumn{1}{l|}{FKG}     & 21.08        & 14.61      & 25.57        & 27.83       \\
\multicolumn{1}{l|}{MSDP-KG (ran.)}   & 8.73 &  8.56  & 15.35     & 16.37       \\
\multicolumn{1}{l|}{MSDP-KG (ppl.)}      & 9.61             & 9.48           & 16.95          & 17.83       \\
\multicolumn{1}{l|}{MSDP-KG (que.)}   & \textbf{23.68}        & \textbf{15.93}      & \textbf{27.88}        & \textbf{31.55}       \\ \bottomrule
\multicolumn{5}{c}{\textit{\textbf{Wizard of Wikipedia (Unseen)}}}   \\ \midrule
\multicolumn{1}{l|}{FKG}      & 9.01         & 8.26       & 15.61        & 16.07       \\
\multicolumn{1}{l|}{MSDP-KG (ran.)}   & 8.89   & 9.11    &  16.19    & 16.42       \\
\multicolumn{1}{l|}{MSDP-KG (ppl.)}      & 9.94     &  10.08    & 17.91   & 18.44       \\
\multicolumn{1}{l|}{MSDP-KG (que.)}    & \textbf{11.54}         & \textbf{10.53}      & \textbf{19.05}        & \textbf{20.15}       \\ \bottomrule
\end{tabular}
\end{adjustbox}
\caption{Ablation study for knowledge generation models. ``ran.'' denotes the prompts are randomly selected, ``ppl.'' denotes the prompts are selected based on the lowest perplexity, and ``que.'' denotes the prompts are selected based on the query.}
\label{tab:ablationKnowledgeGen}
\end{table}

\begin{table}[]
\renewcommand{\arraystretch}{1.1}
\centering
\begin{adjustbox}{width={0.49\textwidth},totalheight={\textheight},keepaspectratio}
\begin{tabular}{lccccc}
\toprule
\multicolumn{1}{l|}{\textbf{Models}}                       & \textbf{B}     & \textbf{M}    & \textbf{R-L}   & \textbf{F1}    & \textbf{KF1}   \\ \bottomrule
\multicolumn{6}{c}{\textit{\textbf{Wizard of Wikipedia (Seen)}}}                                \\ \midrule
\multicolumn{1}{l|}{FCM w/ FKG}             & 8.97           & 8.67          & 15.36          & 18.31          & 18.85          \\
\multicolumn{1}{l|}{FCM w/ MSDP-KG (ppl.)} & 6.93           & 7.67          & 14.01          & 16.89          & 13.59          \\
\multicolumn{1}{l|}{FCM w/ MSDP-KG (que.)} & \textbf{10.17} & 9.34          & 16.60          & \textbf{19.45} & 21.02          \\
\multicolumn{1}{l|}{MSDP (ppl.)}      & 8.18           & 8.43          & 17.46          & 15.92          & 14.73          \\
\multicolumn{1}{l|}{MSDP (que.)}      & 9.97           & \textbf{9.95} & \textbf{18.62} & 17.57          & \textbf{22.95} \\ \bottomrule
\multicolumn{6}{c}{\textit{\textbf{Wizard of Wikipedia (Unseen)}}}      \\ \midrule
\multicolumn{1}{l|}{FCM w/ FKG}             & 6.73           & 7.19          & 12.97          & 14.68          & 12.59          \\
\multicolumn{1}{l|}{FCM w/ MSDP-KG (ppl.)} & 7.03           & 7.58          & 13.81          & 16.54          & 13.23          \\
\multicolumn{1}{l|}{FCM w/ MSDP-KG (que.)} & 7.12           & 7.70          & 13.93          & \textbf{16.75} & 13.96          \\
\multicolumn{1}{l|}{MSDP (ppl.)}      & 7.95           & 8.46          & 17.14          & 15.56          & 15.49          \\
\multicolumn{1}{l|}{MSDP (que.)}      & \textbf{8.30}  & \textbf{8.65} & \textbf{17.40} & 16.00          & \textbf{16.57} \\ \bottomrule
\end{tabular}
\end{adjustbox}
\caption{Ablation study for knowledgeable conversational models. ``MSDP (ppl.)'' and ``MSDP (que.)'' uses ``MSDP-KG (ppl.)'' and ``MSDP-KG (que.)'', respectively, as the knowledge generator.}
\label{tab:ablationDialog}
\end{table}

The ablation studies between perplexity-based sample selection and query-based sample selection are shown in Table~\ref{tab:ablationKnowledgeGen} and Table~\ref{tab:ablationDialog}. We also add finetuning-based knowledge generation (FKG), and sample selection by random into the comparison to better analyze the perplexity-based sample selection method. 

\paragraph{Knowledge Generation}
From Table~\ref{tab:ablationKnowledgeGen}, we can see that perplexity-based selection generally achieves better results across all automatic metrics compared to the sample selection by random, which confirms the effectiveness of using easy to understand samples for prompting. We find that MSDP-KG (ppl.) performs much worse than FKG in WoW (seen). It is because FKG fully utilize the knowledge information from the database which covers all the topics in WoW (seen), but MSDP-KG (ppl.) just uses 10 samples from the database. However, MSDP-KG (ppl.) can outperform FKG in WoW (unseen), which illustrates the generalization ability of perplexity-based selection. Query-based sample selection can remarkably outperform the perplexity-based sample selection on all metrics. It shows that using similar samples to the current conversation is a more effective approach than using fixed samples for all inputs.

\paragraph{Response Generation}
As shown in Table~\ref{tab:ablationDialog}, we can see that better knowledge generation methods generally bring better response generations.
Dialogue models using MSDP-KG (que.) as the knowledge generator generally outperforms the ones using MSDP-KG (ppl.) as the knowledge generator.
Similar to what we have observed in the knowledge generation, ``FCM w/ FKG'' outperforms ``FCM w/ MSDP-KG (ppl.)'' in WoW (seen), since FKG fully uses the samples in the database. However, ``FCM w/ MSDP-KG (ppl.)'' can surpass ``FCM w/ FKG'' in WoW (unseen) due to a better generalization ability of MSDP-KG (ppl.).

\section{Model Scaling Results} \label{appendixC}
The automatic metrics for knowledge generation and response generation in terms of different model sizes are shown in Table~\ref{tab:ablationKnowledgeGenOverSize} and Table~\ref{tab:ablationDialogSize}. 
We observe that when the model sizes are comparable, MSDP is able to achieve comparable or even better results than the ``FCM w/ MSDP-KG''. 
Moreover, we find that larger LMs generally bring better results across all metrics for both knowledge generation and response generation. Furthermore, the 530b LM significantly improves the results across metrics for WoW (unseen), which confirms the strong generation ability of the 530B LM. The relatively small improvement made by the 530B LM in WoW (seen) is because MSDP (1.3b) has already achieved good performance, making it more difficult to improve upon it.

\section{Generation Examples} \label{appendixD}

We provide a few generation examples for FCM w/ DPR (wiki), MSDP (357m), MSDP (1.3b), and MSDP (530b) (shown in Table~\ref{tab:example1},~\ref{tab:example2}, and~\ref{tab:example3}). The samples are selected from WoW (unseen) and WoI.

\begin{table}[]
\renewcommand{\arraystretch}{1.1}
\centering
\begin{adjustbox}{width={0.42\textwidth},totalheight={\textheight},keepaspectratio}
\begin{tabular}{lcccc}
\toprule
\multicolumn{1}{l|}{\textbf{Models}}  & \textbf{B}     & \textbf{M}     & \textbf{R-L}   & \textbf{F1}    \\ \bottomrule
\multicolumn{5}{c}{\textit{\textbf{Wizard of Wikipedia (Seen)}}}                                          \\ \midrule
\multicolumn{1}{l|}{MSDP-KG (126m)} & 23.68          & 15.93          & 27.88          & 31.55          \\
\multicolumn{1}{l|}{MSDP-KG (357m)} & 24.48          & 16.37          & 28.74          & 33.16          \\
\multicolumn{1}{l|}{MSDP-KG (1.3b)} & 25.62 & 17.18 & 29.66 & 34.52 \\
\multicolumn{1}{l|}{MSDP-KG (530b)} & \textbf{27.45} & \textbf{19.34} & \textbf{33.09} & \textbf{35.73} \\
\bottomrule
\multicolumn{5}{c}{\textit{\textbf{Wizard of Wikipedia (Unseen)}}}             \\ \midrule
\multicolumn{1}{l|}{MSDP-KG (126m)} & 11.54          & 10.53          & 19.05          & 20.15          \\
\multicolumn{1}{l|}{MSDP-KG (357m)} & 12.38          & 11.10          & 19.64          & 21.98          \\
\multicolumn{1}{l|}{MSDP-KG (1.3b)} & 13.49 & 11.94 & 20.68 & 23.65 \\
\multicolumn{1}{l|}{MSDP-KG (530b)} & \textbf{18.50} & \textbf{15.15} & \textbf{25.87} & \textbf{29.40} \\
\bottomrule
\end{tabular}
\end{adjustbox}
\caption{Ablation study for MSDP-KG (que.) on different model sizes.}
\label{tab:ablationKnowledgeGenOverSize}
\end{table}

\section{Human Evaluation} \label{appendixE}
\subsection{Human Evaluation Setup}
Both knowledge generation and response generation are evaluated on Amazon Mechanical Turk (AMT). We set up all evaluations as independent AMT tasks to ensure the tasks do not influence each other. To reduce the noise in our labeling process, we only accepted workers with an approval rating over 95\% and who have over 1k accepted jobs. 
Each worker was asked to annotate 10 cases at a time, and we added one control case (very easy to annotate) among them. If a worker provides the wrong judgement for the control case, their annotations were discarded. 
We randomly sample 90 cases for each model in each dataset, and then calculate the averaged score for each metric.

\begin{table}[]
\renewcommand{\arraystretch}{1.1}
\centering
\begin{adjustbox}{width={0.49\textwidth},totalheight={\textheight},keepaspectratio}
\begin{tabular}{lccccc}
\toprule
\multicolumn{1}{l|}{\textbf{Models}}         & \textbf{B}     & \textbf{M}     & \textbf{R-L}   & \textbf{F1}    & \textbf{KF1}   \\ \bottomrule
\multicolumn{6}{c}{\textit{\textbf{Wizard of Wikipedia (Seen)}}}       \\ \midrule
\multicolumn{1}{l|}{FCM w/ MSDP-KG (126m)}  & 10.17          & 9.34           & 16.60          & 19.45          & 21.02          \\
\multicolumn{1}{l|}{FCM w/ MSDP-KG (357m)}  & 10.27          & 9.45           & 16.62          & 20.03          & 21.68          \\
\multicolumn{1}{l|}{FCM w/ MSDP-KG (1.3b)}  & 10.49 & 9.60    & 16.93    & 20.39 & 22.35   \\
\multicolumn{1}{l|}{MSDP (357m)} & 9.97           & 9.95           & 18.62          & 17.57          & 22.95          \\
\multicolumn{1}{l|}{MSDP (1.3b)} & 10.47  & 11.13 & 19.88 & 19.13      & 29.30 \\ 
\multicolumn{1}{l|}{MSDP (530b)} & \textbf{10.83} &  \textbf{12.17} & \textbf{20.35} & \textbf{20.45}          & \textbf{30.38} \\ \bottomrule
\multicolumn{6}{c}{\textit{\textbf{Wizard of Wikipedia (Unseen)}}}       \\ \midrule
\multicolumn{1}{l|}{FCM w/ MSDP-KG (126m)}  & 7.12           & 7.70           & 13.93          & 16.75          & 13.96          \\
\multicolumn{1}{l|}{FCM w/ MSDP-KG (357m)}  & 7.25           & 7.80           & 14.03          & 16.93          & 14.78          \\
\multicolumn{1}{l|}{FCM w/ MSDP-KG (1.3b)}  & 7.64           & 8.07           & 14.46          & 17.57 & 15.98          \\
\multicolumn{1}{l|}{MSDP (357m)} & 8.30           & 8.65           & 17.40          & 16.00          & 16.57          \\ 
\multicolumn{1}{l|}{MSDP (1.3b)} & 8.84  & 9.16  & 18.10 & 17.03          & 20.39 \\ 
\multicolumn{1}{l|}{MSDP (530b)} & \textbf{9.54}  &  \textbf{11.47} & \textbf{19.26} &  \textbf{18.73}  &  \textbf{25.39}  \\ \bottomrule
\end{tabular}
\end{adjustbox}
\caption{Ablation study for knowledgeable conversational models on different model sizes.}
\label{tab:ablationDialogSize}
\end{table}

\subsection{Human Evaluation Interface}
\begin{figure*}[t!]
\centering
\resizebox{0.99\textwidth}{!}{
    \includegraphics{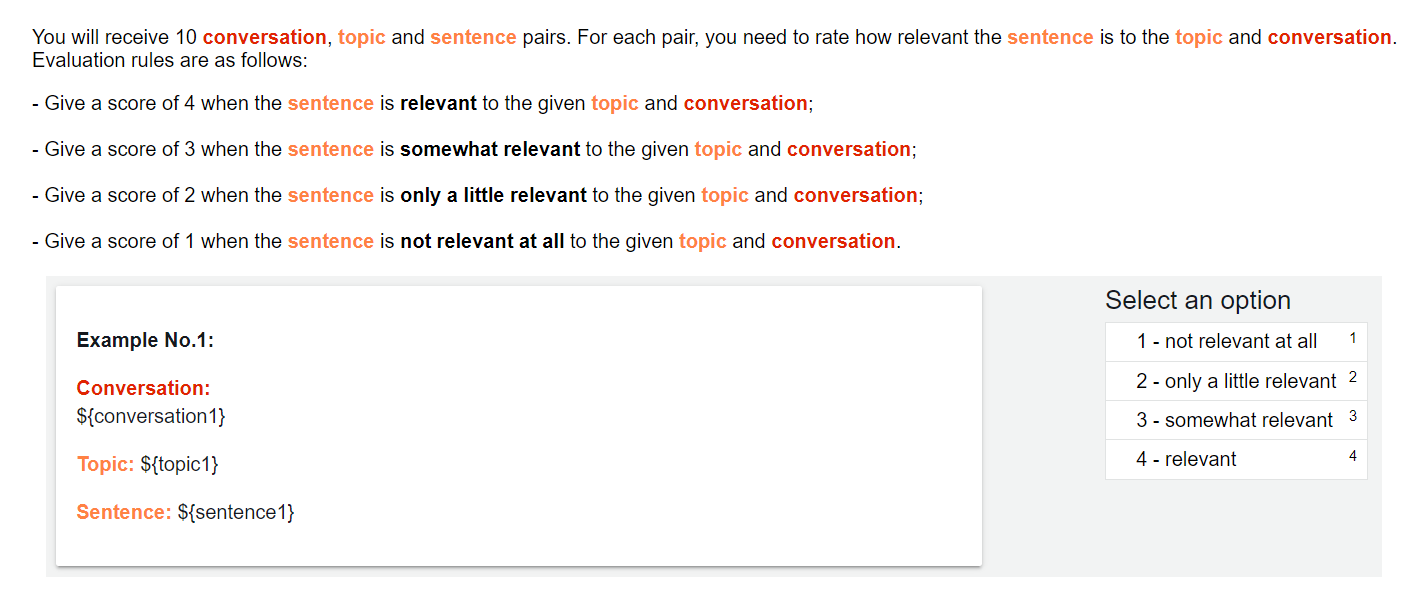}
}
\caption{Knowledge relevance. Note that there are 10 examples in total for one batch. Since all examples follow the same template, we just put one example to avoid the redundancy in these Figure (Same for others).}
\label{fig:knowledgerelevance}
\end{figure*}

\begin{figure*}[t!]
\centering
\resizebox{0.99\textwidth}{!}{
    \includegraphics{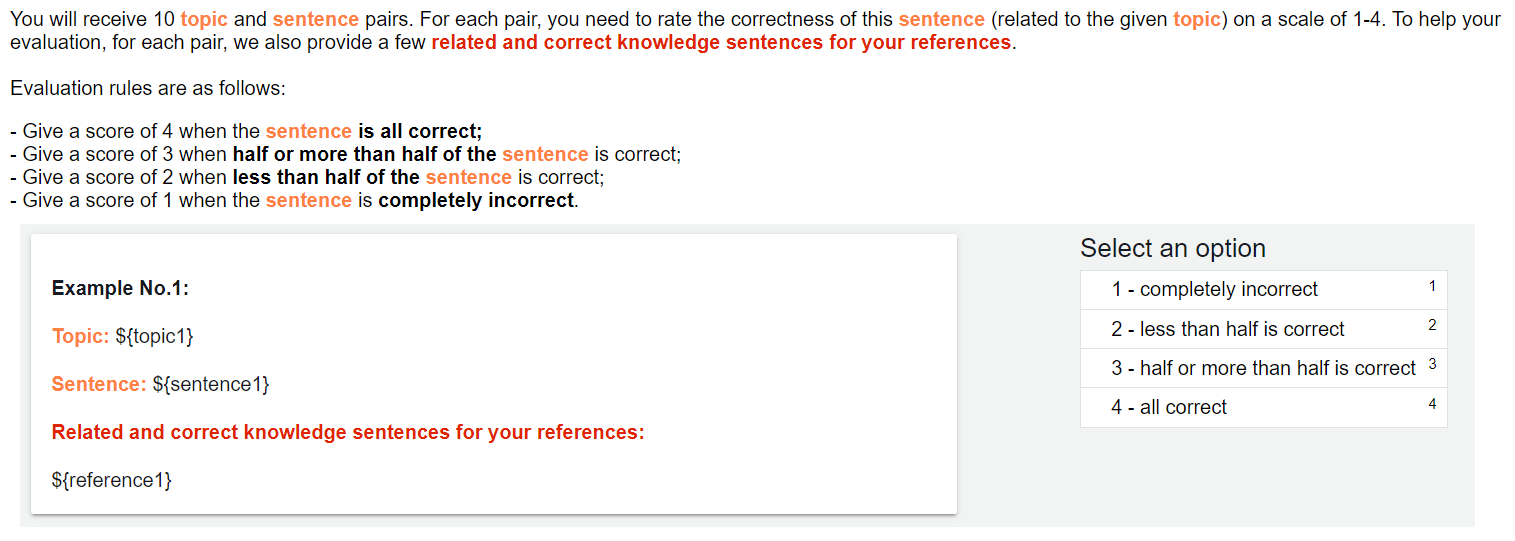}
}
\caption{Knowledge correctness.}
\label{fig:knowledgecorrectness}
\end{figure*}

\begin{figure*}[t!]
\centering
\resizebox{0.99\textwidth}{!}{
    \includegraphics{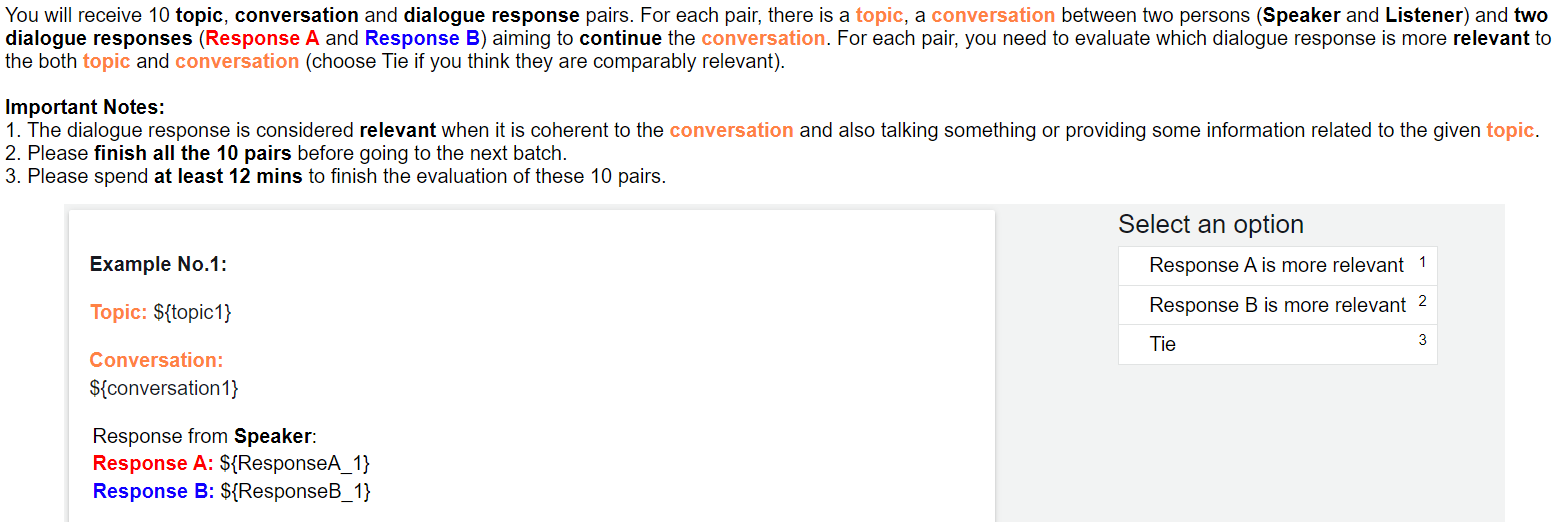}
}
\caption{Response relevance.}
\label{fig:responserelevance}
\end{figure*}

\begin{figure*}[t!]
\centering
\resizebox{0.99\textwidth}{!}{
    \includegraphics{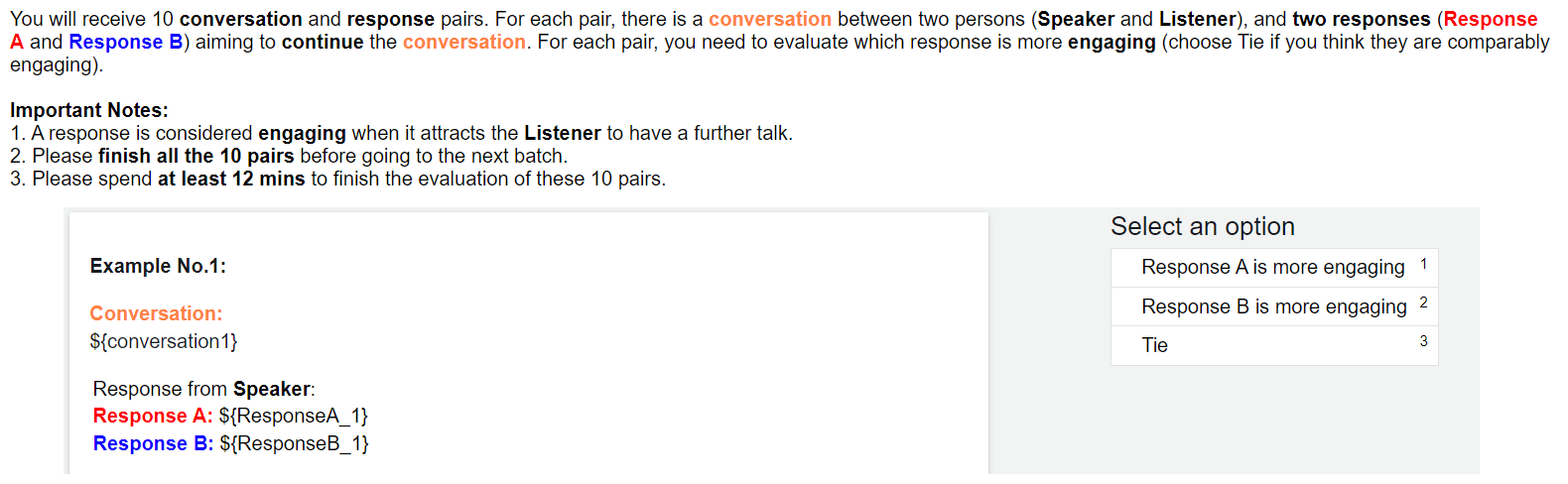}
}
\caption{Response engagement.}
\label{fig:responseengagement}
\end{figure*}

\begin{figure*}[t!]
\centering
\resizebox{0.99\textwidth}{!}{
    \includegraphics{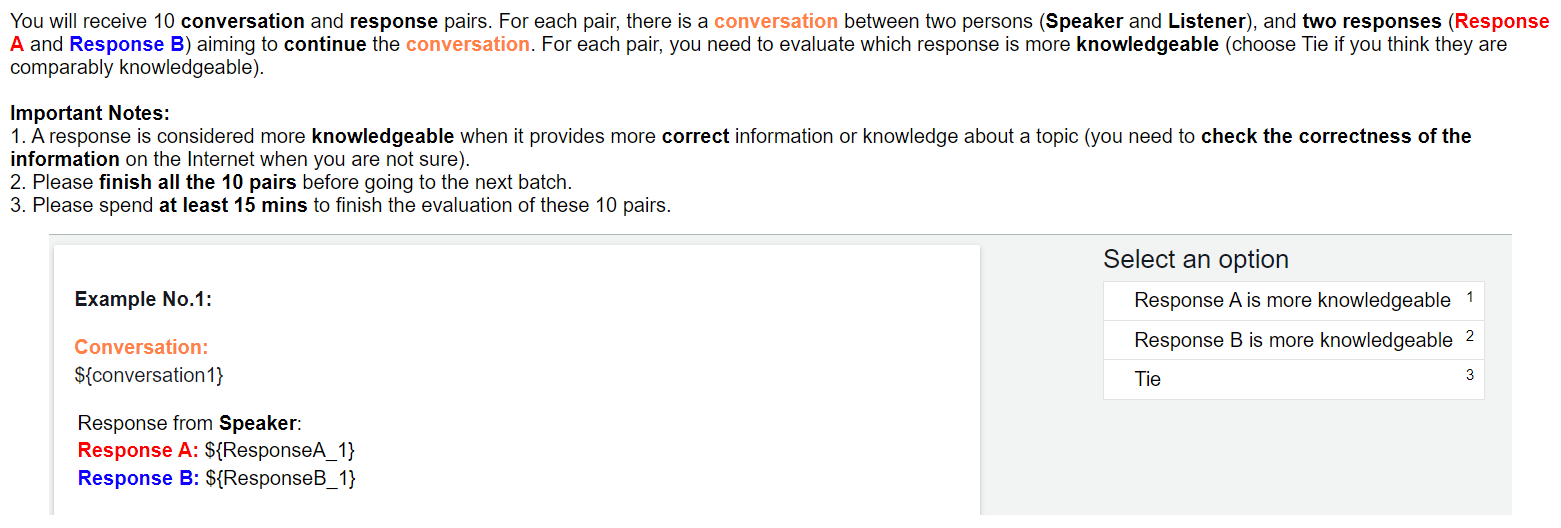}
}
\caption{Response knowledgeability.}
\label{fig:responseknowledgeability}
\end{figure*}

We provide the interfaces used for human
evaluations, which are shown from Figure~\ref{fig:knowledgerelevance} to Figure~\ref{fig:responseknowledgeability}.

\section{Details of Finetuning DPR} \label{appendixF}

\subsection{Overview of DPR}
Dense passage retriever (DPR)~\cite{karpukhin2020dense} uses a dense passage encoder $E_P(\cdot)$ which maps any text passage to a d-dimensional real-valued vectors and builds an index for all the passages that we will use for retrieval.
At run-time, DPR applies a different encoder (question encoder), $E_Q(\cdot)$, that maps the input question to a d-dimensional vector, and retrieves the passages of which vectors are the closest to the question vector.
The similarity between the question and the passage is based on the dot product of their vectors.

\subsection{Finetuning on Dialogue Scenario}
DPR is originally pretrained based on the QA dataset with the Wikipedia as the knowledge source. Since there is a discrepancy between the dialogue domain and the QA domain, it could make the retrieval ability of DPR not optimal for the dialogue scenario. 
Therefore, we attempt to construct a stronger baseline by finetuning DPR on the dialogue scenario using the training dataset of Wizard of Wikipedia (WoW)~\cite{dinan2018wizard}.

Concretely, we further finetune DPR in the dialogue scenario by following its original training procedure, and maximize the dot product similarity between the dialog example ($d_i$) and the corresponding ground truth knowledge ($k_i$):
\begin{equation*}
    \text{sim}(d_i, k_i) = E_Q(t_i+h_i)^\intercal E_P(k_i),
\end{equation*}
where $d_i$ and $k_i$ are training samples in $D$ (training dataset of WoW), and $d_i$ is a concatenation of the topic $t_i$ and dialogue history $h_i$.

\section{Discussion on Baseline Selection} \label{appendixG}
Although we used several baselines for comparisons with our methods, we did not consider baselines that incorporate the ground truth knowledge information for the response generation.
Such baselines~\cite{kim2019sequential,zhao2019low,zhao2020knowledge,zhan2021augmenting} retrieve knowledge from a small set of candidates (about 7 examples) that are relevant to the dialogue history, and one of the candidates is the ground truth knowledge. In contrast, our model (MSDP) does not leverage such ground truth information, since it is usually not available in real world scenarios. Therefore, we did not compare our model with those baselines in our experiments.

In addition, given that our model does not need any fine-tuning and uses only 20 samples as prompts for the response generation, FCM w/ DPR and FCM w/ FKG makes it a strong baseline for our model to compare with, since it is finetuned on the entire training dataset.


\section{Analysis on Response Generation} \label{appendixH}
As we can see from the provided generation examples (shown in Table~\ref{tab:example1}, \ref{tab:example2}, and \ref{tab:example3}), our model (MSDP) is able to effectively leverage the generated knowledge in the first-stage prompting. In this part, we further analyze how much our model tends to copy the generated knowledge and our model generates response when the generated knowledge is not very relevant to the current conversation.

We quantify the proportion of the knowledge in the generated responses, which we formulate as follows:
\begin{equation}
    ratio_{knwl} = \frac{\texttt{\# \{overlap tokens\}}}{\texttt{\# \{response tokens\}}},
\end{equation}
Where \texttt{\# \{overlap tokens\}} denotes the number of overlap tokens between the generated knowledge and the generated response, and \texttt{\# \{response tokens\}} denotes the number of tokens in the response.
The ratios for MSDP (357m), MSDP (1.3b), and MSDP (530b) in the WoW (unseen) are 49.67\%, 46.11\%, and 44.19\%, respectively. This suggests that the response is not just simply copies of the knowledge, it also contains additional information to ensure the relevance and engagingness. 

We further analyze the cases where the generated knowledge is not very relevant to the conversation. As we can see from Table~\ref{tab:case1}, our model could try to make the conversation coherent and engaging by leveraging the generated knowledge although it is not very relevant to the current conversation\footnote{MSDP (530b) generates relevant knowledge in this example. Hence, we do not put the example from MSDP (530b) in this Table.}. Additionally, from Table~\ref{tab:case2}, our model could choose not to copy the knowledge and manage to generate response only based on the dialogue context, so as to ensure the relevance of the response.

\begin{table}[t!]
\renewcommand{\arraystretch}{1.2}
\centering
\begin{adjustbox}{width={0.49\textwidth},totalheight={\textheight},keepaspectratio}
\begin{tabular}{llll}
\toprule
\multicolumn{4}{l}{\textit{\textbf{Conversation:}}}        \\ 
\multicolumn{4}{l}{- Do you know what wavelengths of light purple occurs at?}   \\ \midrule
\multicolumn{4}{l}{\textit{\textbf{Knowledge from MSDP (357m):}}}    \\
\multicolumn{4}{l}{Purple is a composite color made by combining red and blue.}       \\ \midrule
\multicolumn{4}{l}{\textit{\textbf{Response from MSDP (357m):}}}      \\ 
\multicolumn{4}{l}{\begin{tabular}[c]{@{}l@{}}I'm not sure about that , but I know that purple is a composite \\ color made by combining red and blue .\end{tabular}} \\ \bottomrule
\end{tabular}
\end{adjustbox}
\caption{Example No.1 where the generated knowledge is not very relevant the current conversation.}
\label{tab:case1}
\end{table}

\begin{table}[t!]
\renewcommand{\arraystretch}{1.2}
\centering
\begin{adjustbox}{width={0.49\textwidth},totalheight={\textheight},keepaspectratio}
\begin{tabular}{llll}
\toprule
\multicolumn{4}{l}{\textit{\textbf{Conversation:}}}    \\
\multicolumn{4}{l}{\begin{tabular}[c]{@{}l@{}}- Skiing is simply using two poles and skiboards to move quickly \\ down ice, skiing has a five millennia history. \\ - It looks easy when I see people skiing on TV, but I would be the \\ first to fall right off the bat! Have you actually skied before? \end{tabular}} \\ \midrule
\multicolumn{4}{l}{\textit{\textbf{Knowledge from MSDP (357m):}}}   \\
\multicolumn{4}{l}{\begin{tabular}[c]{@{}l@{}}Skiing is a sport in which a skier skis down a slope, usually in a \\ snow-covered area, at high speeds.\end{tabular}}       \\
\multicolumn{4}{l}{\textit{\textbf{Response from MSDP (357m):}}}    \\
\multicolumn{4}{l}{I have skied before . I found it interesting .}   \\ \midrule
\multicolumn{4}{l}{\textit{\textbf{Knowledge from MSDP (530b):}}}   \\
\multicolumn{4}{l}{\begin{tabular}[c]{@{}l@{}}Skiing is a sport and recreational activity which involves descending \\ a slope covered with snow on skis with fixed-heel bindings.\end{tabular}}       \\
\multicolumn{4}{l}{\textit{\textbf{Response from MSDP (530b):}}}    \\
\multicolumn{4}{l}{\begin{tabular}[c]{@{}l@{}}Yeah i have , and you are right it is really easy when you see people \\ skiing on TV , but it really is not that easy ! It does not look easy \\ when you actually try it yourself .\end{tabular}}   \\ \bottomrule
\end{tabular}
\end{adjustbox}
\caption{Example No. 2 where the generated knowledge is not very relevant for the conversation.}
\label{tab:case2}
\end{table}

\section{Discussion on Prompt Design} \label{appendixI}
We have explored different prompt designs for the response generation, which is shown in Table~\ref{tab:promptdesign}. We find that comprehensive connection tokens (e.g., \textcolor{red}{Systems:}, \textcolor{red}{Users:}, \textcolor{red}{We know that:}) are important for our model to understand how to generate the expected response. Without them (shown in \textit{\textbf{Prompt Format 1}}), the performance drops.
In addition, the position of <knowledge> also matters. If we put the <knowledge> ahead of the dialogue history (shown in \textit{\textbf{Prompt Format 2}}), the performance also drops.

\begin{table*}[]
\renewcommand{\arraystretch}{1.1}
\centering
\begin{adjustbox}{width={0.99\textwidth},totalheight={\textheight},keepaspectratio}
\begin{tabular}{l}
\toprule
\textit{\textbf{Prompt Format 1:}}       \\
\begin{tabular}[c]{@{}l@{}}\textless{}topic 1\textgreater ~ \textless{}dialogue history 1\textgreater ~ \textless{}knowledge 1\textgreater ~ =\textgreater ~ \textless{} response 1\textgreater\\ … …\\ \textless{}topic n\textgreater ~ \textless{}dialogue history n\textgreater ~ \textless{}knowledge n\textgreater ~ =\textgreater ~ \textless{}response n\textgreater\\ \textless{}current topic\textgreater ~ \textless{}current dialogue history\textgreater ~ \textless{}generated knowledge\textgreater ~ =\textgreater ~ \colorbox{green}{\textless{}Expect model to generate\textgreater{}}\end{tabular}        \\ \midrule
\textit{\textbf{Prompt Format 2:}}         \\
\begin{tabular}[c]{@{}l@{}}\textless{}topic 1\textgreater ~ \textcolor{red}{We know that:} \textless{}knowledge 1\textgreater ~ \textcolor{red}{System:} \textless{}system-turn\textgreater ~ \textcolor{red}{User:} \textless{}user-turn\textgreater ~ \textcolor{red}{System replies:} \textless{}response 1\textgreater\\ … …\\ \textless{}topic n\textgreater ~ \textcolor{red}{We know that:} \textless{}knowledge n\textgreater ~ \textcolor{red}{System:} \textless{}system-turn\textgreater ~ \textcolor{red}{User:} \textless{}user-turn\textgreater ~ \textcolor{red}{System replies:} \textless{}response n\textgreater\\ \textless{}current topic\textgreater{} \textcolor{red}{We know that:} \textless{}generated knowledge\textgreater ~ \textcolor{red}{System:} \textless{}s-turn\textgreater ~ \textcolor{red}{User:} \textless{}u-turn\textgreater ~ \textcolor{red}{System replies:} \colorbox{green}{\textless{}Expect model to generate\textgreater{}}\end{tabular} \\ \midrule
\textit{\textbf{Prompt Format 3 (Our final format which gives best performance):}}     \\
\begin{tabular}[c]{@{}l@{}}\textless{} topic 1\textgreater ~ \textcolor{red}{System:} \textless{}system-turn\textgreater ~ \textcolor{red}{User:} \textless{}user-turn\textgreater ~ \textcolor{red}{We know that:} \textless{}knowledge 1\textgreater ~ \textcolor{red}{System replies:} \textless{}response 1\textgreater\\ … …\\ \textless{}topic n\textgreater ~ \textcolor{red}{System:} \textless{}system-turn\textgreater ~ \textcolor{red}{User:} \textless{}user-turn\textgreater ~ \textcolor{red}{We know that:} \textless{}knowledge n\textgreater ~ \textcolor{red}{System replies:} \textless{}response n\textgreater\\ \textless{}current topic\textgreater ~ \textcolor{red}{System:} \textless{}s-turn\textgreater ~ \textcolor{red}{User:} \textless{}u-turn\textgreater ~ \textcolor{red}{We know that:} \textless{}generated knowledge\textgreater ~ \textcolor{red}{System replies:} \colorbox{green}{\textless{}Expect model to generate\textgreater{}}\end{tabular}  \\ \bottomrule 
\end{tabular}
\end{adjustbox}
\caption{Different prompt formats we explore for the response generation. <xxx> can be replaced with the corresponding content in the data sample. The words marked in \textcolor{red}{red color} are the additional connection words to make the prompts comprehensible for models. \colorbox{green}{<Expect model to generate>} denotes the response we expect to get from our model.}
\label{tab:promptdesign}
\end{table*}

\clearpage

\begin{table*}[]
\renewcommand{\arraystretch}{1.1}
\centering
\begin{adjustbox}{width={0.95\textwidth},totalheight={\textheight},keepaspectratio}
\begin{tabular}{ll}
\toprule
Topic:        & inhaling helium                           \\ \cmidrule{2-2}
Conversation: & \begin{tabular}[c]{@{}l@{}}- I bought a tank for my son's birthday party. Needless to say, I've been doing funny \\   voices a lot. Do you think I hurt myself?\\ - If you inhaled straight from the tank, yes! From the balloon is fine.\\ - I did it from the balloon lol. I have trouble inhaling it for some reason. Is there any \\   long-term risks with helium inhalation?\end{tabular} \\ \toprule
\multicolumn{2}{l}{\textit{\textbf{FCM w/ DPR (wiki)}}}       \\
Knowledge     & \begin{tabular}[c]{@{}l@{}}Although babies born at full-term are still susceptible to this, they are likely to \\ respond better and thus, tend to have better outcomes in response to low-oxygen \\ events.\end{tabular}      \\ \cmidrule{2-2}
Response      & Yes , babies born at full-term are still susceptible to this .  \\ \toprule
\multicolumn{2}{l}{\textit{\textbf{MSDP (357m)}}}       \\
Knowledge     & Long-term risks for inhaling helium include: shortness of breath.        \\ \cmidrule{2-2}
Response      & I know that I have never had any problems with helium inhalation .     \\ \toprule
\multicolumn{2}{l}{\textit{\textbf{MSDP (1.3b)}}}      \\
Knowledge     & \begin{tabular}[c]{@{}l@{}}The long-term risks for inhaling helium include: shortness of breath, chest tightness, \\ and coughing.\end{tabular}     \\ \cmidrule{2-2}
Response      & \begin{tabular}[c]{@{}l@{}}Yeah , I know that long term risks for inhaling helium includes shortness of breath, \\ chest tightness, and coughing.\end{tabular}   \\ \toprule
\multicolumn{2}{l}{\textit{\textbf{MSDP (530b)}}}      \\
Knowledge     & \begin{tabular}[c]{@{}l@{}}Long-term inhalation of high concentrations of helium can cause asphyxiation due \\ to oxygen deprivation.\end{tabular}     \\ \cmidrule{2-2}
Response      & \begin{tabular}[c]{@{}l@{}}Yes. Definitely! The biggest danger from inhaling helium is that the oxygen is pushed \\ out of your lungs and causes asphyxiation due to oxygen deprivation.\end{tabular}    \\ \bottomrule               
\end{tabular}
\end{adjustbox}
\caption{Example 1}
\label{tab:example1}
\end{table*}

\begin{table*}[]
\renewcommand{\arraystretch}{1.1}
\centering
\begin{adjustbox}{width={0.95\textwidth},totalheight={\textheight},keepaspectratio}
\begin{tabular}{ll}
\toprule
Topic:        & Kyoto      \\ \cmidrule{2-2}
Conversation: & \begin{tabular}[c]{@{}l@{}}- Good morning, this is my first day visiting japan.\\ - Welcome! You should visit Ancient Kyoto.\\ - I've since kyoto in many animes and would love to see it in person\end{tabular} \\ \toprule
\multicolumn{2}{l}{\textit{\textbf{FCM w/ DPR (wiki)}}}        \\
Knowledge     & In "Onisarashi-hen", Natsumi Kimiyoshi is the visitor of Okinomiya.          \\ \cmidrule{2-2}
Response      & I 'm sure you 'll enjoy it . In the anime , Natsumi Kimiyoshi is the visitor of Okinawa .   \\ \toprule
\multicolumn{2}{l}{\textit{\textbf{MSDP (357m)}}}      \\
Knowledge     & Kyoto is considered the cultural capital of Japan.         \\ \cmidrule{2-2}
Response      & Great ! I remember Kyoto is considered the cultural capital of Japan .      \\ \toprule
\multicolumn{2}{l}{\textit{\textbf{MSDP (1.3b)}}}     \\
Knowledge     & Kyoto is considered the cultural capital of Japan and a major tourist destination.        \\ \cmidrule{2-2}
Response      & \begin{tabular}[c]{@{}l@{}}That's nice! Kyoto is considered the cultural capital of Japan and a major tourist \\ destination.\end{tabular}       \\ \toprule
\multicolumn{2}{l}{\textit{\textbf{MSDP (530b)}}}         \\
Knowledge     & Kyoto has many UNESCO World Heritage Sites       \\ \cmidrule{2-2}
Response      & \begin{tabular}[c]{@{}l@{}}Oh it is nice ! you should definitely go ! Kyoto has many UNESCO World Heritage \\ Sites which you should visit .\end{tabular}      \\ \bottomrule                     
\end{tabular}
\end{adjustbox}
\caption{Example 2}
\label{tab:example2}
\end{table*}

\begin{table*}[]
\renewcommand{\arraystretch}{1.1}
\centering
\begin{adjustbox}{width={0.95\textwidth},totalheight={\textheight},keepaspectratio}
\begin{tabular}{ll}
\toprule
Topic:        & Online shopping    \\ \cmidrule{2-2}
Conversation: & - I love using Amazon , have you tried it ?       \\ \toprule
\multicolumn{2}{l}{\textit{\textbf{FCM w/ DPR (wiki)}}}          \\
Knowledge     & This would now be referred to as online shopping.    \\ \cmidrule{2-2}
Response      & Yes , I have tried it . I like that it is now referred to as online shopping .   \\ \toprule
\multicolumn{2}{l}{\textit{\textbf{MSDP (357m)}}}          \\
Knowledge     & \begin{tabular}[c]{@{}l@{}}Online shopping is the process of purchasing goods or services online from a \\ website or other online store.\end{tabular}     \\ \cmidrule{2-2}
Response      & \begin{tabular}[c]{@{}l@{}}Yes I have . I love using Amazon . I know that Online shopping is the process of \\ purchasing goods or services from a website or other online service provider .\end{tabular}    \\ \toprule
\multicolumn{2}{l}{\textit{\textbf{MSDP (1.3b)}}}        \\
Knowledge     & Online shopping is the use of the Internet to purchase goods and services.       \\ \cmidrule{2-2}
Response      & \begin{tabular}[c]{@{}l@{}}Yes, I love it . I know that online shopping is the use of the Internet to purchase \\ goods and services .\end{tabular}              \\ \toprule
\multicolumn{2}{l}{\textit{\textbf{MSDP (530b)}}}             \\
Knowledge     & \begin{tabular}[c]{@{}l@{}}Online shopping is a form of electronic commerce which allows consumers to directly \\ buy goods or services from a seller over the Internet using a web browser.\end{tabular}     \\ \cmidrule{2-2}
Response      & \begin{tabular}[c]{@{}l@{}}I have . I really love it . It is very convenient since it allows consumers to directly buy \\ goods or services from a seller over the Internet using a web browser.\end{tabular} \\ \bottomrule
\end{tabular}
\end{adjustbox}
\caption{Example 3}
\label{tab:example3}
\end{table*}

\clearpage

\end{document}